\documentclass{article}

\usepackage{amsmath,amssymb,amsfonts, amsthm}
\usepackage{booktabs}
\usepackage{graphicx}
\usepackage{textcomp}
\usepackage{float} 
\usepackage[table]{xcolor}
\usepackage{color}
\usepackage{url}
\usepackage{microtype}
\usepackage{wrapfig}
\usepackage{bm,bbm}
\usepackage{caption}
\usepackage{subcaption}
\usepackage{mathtools}
\usepackage{graphicx}
\usepackage[export]{adjustbox}

\usepackage{mathtools}
\usepackage{enumitem}

 \usepackage{multirow} 

\newcommand{\mD}{\mathcal{D}}
\newcommand{\real}{\mathbb{R}}

\newcommand{\mG}{\mathcal{G}}

\newcommand{\mT}{\mathcal{T}}
\newcommand{\mC}{\mathcal{C}}
\newcommand{\mL}{\mathcal{L}}
\newcommand{\bx}{\mathbf{x}}
\newcommand{\by}{\mathbf{y}}
\newcommand{\br}{\mathbf{r}}

\usepackage{hyperref}

\usepackage[accepted]{icml2023}

\usepackage{amsmath}
\usepackage{amssymb}
\usepackage{mathtools}
\usepackage{amsthm}

\usepackage[capitalize,noabbrev]{cleveref}

\theoremstyle{plain}

\theoremstyle{definition}

\theoremstyle{remark}

\usepackage[textsize=tiny]{todonotes}

\icmltitlerunning{GraphPPD: Posterior Predictive Modelling for Graph-Level Inference}

\begin{document}

\twocolumn[
\icmltitle{GraphPPD: Posterior Predictive Modelling for Graph-Level Inference}



\icmlsetsymbol{equal}{*}

\begin{icmlauthorlist}
\icmlauthor{Soumyasundar Pal}{huawei}
\icmlauthor{Liheng Ma}{huawei,mcgill,mila}
\icmlauthor{Amine Natik}{huawei,mila,udem}
\icmlauthor{Yingxue Zhang}{huawei}
\icmlauthor{Mark Coates}{mcgill,mila}
\end{icmlauthorlist}

\icmlaffiliation{udem}{Université de Montréal}
\icmlaffiliation{mcgill}{McGill University}
\icmlaffiliation{mila}{Mila - Quebec AI Institute}
\icmlaffiliation{huawei}{Huawei Noah’s Ark Lab}

\icmlcorrespondingauthor{Soumyasundar Pal}{soumyasundar.pal@huawei.com}

\icmlkeywords{Graph classification, variational inference, meta-learning}

\vskip 0.3in
]



\printAffiliationsAndNotice{}  

\begin{abstract}
Accurate modelling and quantification of predictive uncertainty is crucial in deep learning since it allows a model to make safer decisions when the data is ambiguous and facilitates the users' understanding of the model's confidence in its predictions.
Along with the tremendously increasing research focus on \emph{graph neural networks} (GNNs) in recent years, there have been numerous techniques which strive to capture the uncertainty in their predictions. However, most of these approaches are specifically designed for node or link-level tasks and cannot be directly applied to graph-level learning problems.
In this paper, we propose a novel variational modelling framework for the \emph{posterior predictive distribution}~(PPD) to obtain uncertainty-aware prediction in graph-level learning tasks.
Based on a graph-level embedding derived from one of the existing GNNs, our framework can learn the PPD in a data-adaptive fashion.
Experimental results on several benchmark datasets exhibit the effectiveness of our approach.
\end{abstract}

\section{Introduction}
\label{sec:intro}

Recently, \emph{graph neural networks} (GNNs) have become the \emph{de-facto} tool to analyze and learn from network structured data, which is pervasive in neuroscience~\cite{griffa2017transient}, chemistry~\cite{duvenaud2015, gilmer2017neural}, recommendation~\cite{monti2017geometric, ying2018graph}, and social sciences~\cite{kipf2017, monti2019fake}. 
Several avenues have been explored for  designing more expressive GNN architectures~\cite{xu2019, morris2019weisfeiler}. 
Despite their success, most of these approaches suffer from the major weakness that they cannot gauge the uncertainty associated with their predictions. Assessment of predictive uncertainty is crucial in many practical applications if a decision is to be made using the predictions.

In order to address this drawback, there have been research efforts to equip Graph Neural Networks with uncertainty characterization. However, most of these techniques are tailored to node classification~\cite{zhang2019,ma2019,elinas2020}), link prediction~\cite{pal2020,opolka2022}, and recommendation~\cite{sun2020} and cannot handle graph-level learning tasks such as those appearing in~\cite{xu2019, hu2021OpenGraphBenchmark, dwivedi2020benchmarkgnns, morris2020tudataset, yang2022_LS_prediction}. Graph-level inference has diverse applications in molecular property understanding~\cite{hu2021OpenGraphBenchmark}, circuit performance prediction~\cite{hakhamaneshi2022pretraining}, protein structure inference~\cite{xu2019}, and many other fields. Existing techniques from the conventional Bayesian deep learning~\cite{gal2016,li2016} literature are either ineffective for quantifying uncertainty or computationally costly when applied to model the uncertainty associated with the parameters of GNNs.

In this paper, we consider amortized approximation of the posterior predictive density (PPD). We employ a variational inference formulation for tackling graph-level learning tasks in a supervised setting. This framework permits a principled utilization of the training set of graphs and their labels during the inference stage when forming predictions on the test set graphs. Our design of the PPD module is based on a cross-attention mechanism between the embeddings of the query graphs and those of selected training set graphs with known labels. This allows for a direct comparison between different labeled graphs during training, which leads to improved learning of a generalizable PPD module. Comparison with labelled training graphs during the inference stage ameliorates the prediction performance. We evaluate our designed GraphPPD module on 18 commonly used graph classification and regression tasks, as well as on a selective prediction~\cite{kivlichan2021} task to validate the effectiveness of its predictive uncertainty.

The novel contributions in this paper are as follows:\\
1) We propose a novel variational inference based PPD learning framework for application in graph-level learning tasks. \\
2) Our framework can incorporate existing GNNs as graph encoders and can be extended to other domains using suitable feature extractor modules.\\
3) In addition to end-to-end training, our approach allows use of pre-trained graph encoders for learning the PPD module with considerably lower training time.\\
4) Experimental results on diverse benchmark graph-level learning tasks  show that the incorporation of several GNNs and a graph transformer for graph-level representation learning inside our framework proves beneficial in most cases. Besides, we show our predictive uncertainty can effectively serve as the review criterion in the selective prediction~\cite{kivlichan2021} task.

\section{Preliminary}
\label{sec:prelim}
In a supervised setting, the learning goal is to provide a label $\by'$ for a data instance with features $\bx'$. In presence of a training dataset $\mD_{\mL} = \{(\bx_i, \by_i)\}_{i \in \mL}$, obtained by i.i.d. sampling from the same underlying data distribution $p_{true}(\bx, \by)$ as that of $(\bx', \by')$, we aim to learn a conditional distribution model $p(\by|\bx; \gamma)$, parameterized by $\gamma$, which is subsequently used for labelling $\bx'$. In the Bayesian setting, the model parameter $\gamma$ is assumed to be random variable and a suitable prior $p(\gamma)$ is adopted. In order to account for parameter uncertainty while forming the prediction for a new input, the \emph{posterior predictive distribution} (PPD) is evaluated as the expected predictive density w.r.t. to the posterior distribution of $\gamma$, as shown below:
\begin{align}
 p(\by'|\bx', \mD_{\mL}) = \int p(\by'|\bx'; \gamma)p(\gamma|\mD_{\mL})\,d \gamma\,.\label{eq:ppd}    
\end{align}
The training objective is to compute the posterior distribution of $\gamma$, which follows Bayes' rule:
\begin{align}
    p(\gamma|\mD_{\mL}) &\propto p(\gamma) \prod_{i \in \mL} p(\by_i|\bx_i;\gamma)\,.\label{eq:posterior}
\end{align}
If the chosen family of models contains the `true' conditional distribution, i.e., if there exists a $\gamma^*$ such that $p(\by|\bx; \gamma^*) = p_{true}(\by|\bx)$ holds for all $(\bx, \by)$, the PPD results in optimal prediction~\cite{nogales2022}. 

In most practical cases, neither computing the posterior in eq.~\eqref{eq:posterior} nor evaluating the PPD in eq.~\eqref{eq:ppd} can be done analytically. Several techniques for tractable approximation of the posterior $p(\gamma|\mD_{\mL})$ have been proposed based on expectation propagation~\cite{hernandez2015} and  \emph{variational inference} (VI)~\cite{gal2016,sun2017,louizos2017b}. Typically, such approximations are easy to sample from (e.g., multivariate normal distribution with diagonal covariance structure), which allows us to form Monte Carlo estimates of the PPD subsequently. Another class of algorithms resorts to sampling from the posterior via various \emph{Markov Chain
Monte Carlo} (MCMC) methods~\cite{neal1993,korattikara2015,li2016,izmailov2021}. Although these approaches are effective, VI generally introduces a difficult-to-characterize bias in estimating the PPD and underestimates the predictive uncertainty. On the other hand, MCMC  approaches often require prohibitively expensive computation. 
In contrast, recently there has been another line of work~\cite{gordon2018metalearning,garnelo2018b,muller2021}, which attempts to learn an approximation of the PPD from the training data directly using neural networks. 


\section{Problem Statement}
\label{sec:ps}
We address variational modelling of the PPD for graph-level learning tasks, where our goal is to approximate the distribution of the label $\by$ of the input graph $\mG$, conditioned on a set of labelled graphs. 
We consider an inductive setting, where we have access to a labelled dataset $\mD_{\mL} =\{(\mG_i, \by_i)\}_{i \in \mL}$, which is used for model training and subsequently the trained model is used for predicting the labels of the test set graphs $\mG_i$, where $i \in \overline{\mL}$. We assume that $\mD_{\mL}$ is accessible during the inference stage as well. 

As example learning tasks we consider graph classification and graph regression. The graph can be directed or undirected, can have a weighted adjacency matrix, and node and/or edge features associated with it.
The label $\by$ is categorical in a classification setting and real-valued in regression problems. 
The  classification performance is assessed by evaluating accuracy or \emph{area under the ROC (receiver operating characteristic) curve}  ROC-AUC  of the test set predictions. \emph{Mean absolute error} (MAE) and \emph{mean square error} (MSE) are used for regression.


\section{Methodology}
\label{sec:method}

\begin{figure*}
\centering \includegraphics[width=\linewidth, frame]{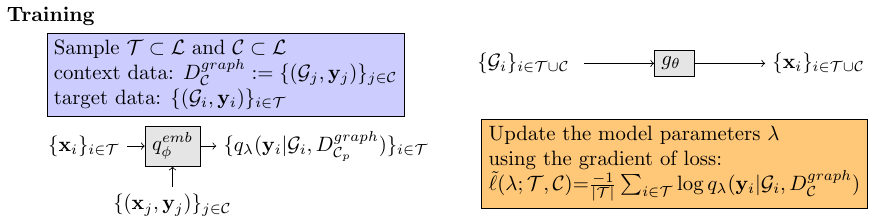} 
\includegraphics[width=\linewidth, frame]{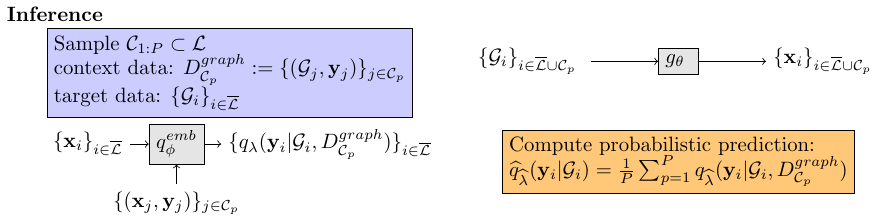}
 \caption{Procedure for GraphPPD training and inference}
\label{fig:graphppd}
\end{figure*}

\subsection{Overall Framework}
We propose a general framework for modelling the \emph{posterior predictive distribution} (PPD) of the labels of a set of unlabelled graphs (targets), conditioned on another set of labelled graphs (contexts). Our architecture is comprised of two components; a) a deterministic feature extractor (graph encoder) module $g_{\theta}(\cdot)$ (e.g. GNNs, graph transformers), which generates a $d_x$-dimensional embedding from an input graph, and b) an amortized PPD approximation module $q_{\phi}^{emb}$, which processes the graph representations of the target and context set, along with the labels of the contexts, to form the approximate posterior predictive distribution. 
We denote the complete set of learnable parameters as $\lambda = \{\theta, \phi\}$. 
During model training, both the targets and the contexts are sampled uniformly from the training data for learning $\lambda$, whereas the test set graphs serve as targets in the inference stage. 
The sampling of targets is innately done by mini-batching in the training process.

Suppose the target and context sets are indexed by $\mT$ and $\mC$, respectively.
To simplify the notation, we abbreviate the context data $\{(\mG_j, \by_j)\}_{j \in \mC}$ in the graph domain by $D_{\mC}^{graph}$. For each graph $\mG_i$, where, $i \in \mT \bigcup \mC$, the embedding $\bx_i=g_{\theta}(\mG_i)$ is obtained by a forward pass through the graph encoder. We use $D_{\mC}^{emb} = \{(\bx_j, \by_j)\}_{j \in \mC}$ to denote the context representations and their labels. $D_{\mC}^{emb}$  is subsequently fed to the amortized PPD module along with the target embeddings $\{\bx_i\}_{i \in \mT}$ for modelling $q_{\lambda}(\{\by_i\}_{i \in \mT}|\{\mG_i\}_{i \in \mT}, D_{\mC}^{graph})$, which is evaluated as:
\begin{align}
q_{\lambda}(\{\by_i\}_{i \in \mT}|\{\mG_i\}_{i \in \mT}, D_{\mC}^{graph})
&=\prod_{i \in T} q_{\lambda}(\by_i|\mG_i, D_{\mC}^{graph})\,,\label{eq:cond_ind}\\
&=\prod_{i \in T} q_{\phi}^{emb}(\by_i|\bx_i,D^{emb}_{\mC}) \,.\label{eq:ppd_graph}
\end{align}

The simplification in eq.~\eqref{eq:cond_ind} follows from the conditional independence of the targets given the contexts. Since there is no stochasticity in the forward pass through $g_{\theta}$ if its parameter $\theta$ is fixed, eq.~\eqref{eq:ppd_graph} follows from eq.~\eqref{eq:cond_ind}.  

\subsection{Posterior predictive distribution module $q_{\phi}^{emb}$}
The criteria for the PPD module are: a) permutation invariance, i.e., prediction for any target should remain the same for any ordering of the contexts; and b) flexibility, i.e, $q_{\phi}^{emb}$ should be able to handle an arbitrary number of targets and contexts. 
Additionally, the design should be expressive such that each target can focus on relevant contexts in an adaptive manner for forming its prediction. 
Inspired by the recent work in \emph{Attentive Neural Processes} (ANPs)~\cite{kim2019}, we employ a cross-attention mechanism between the targets and contexts to design $q_{\phi}^{emb}$.
Specifically, the cross-attention coefficient between the $i$-th  target representation (query) and the $j$-th context representation (key) is computed as : 
\begin{align}
\alpha_{ij} := \text{Softmax}_{j \in \mC} \Bigg(\frac{\langle \mathbf{W}_q \bx_i, \mathbf{W}_k \bx_j \rangle}{\sqrt{d_h}} \Bigg)\,,\label{eq:attn_score} \forall (i,j) \in \mT \times \mC,
\end{align}
where $\mathbf{W}_q, \mathbf{W}_k \in \real^{d_h \times d_x}$ are learnable weight matrices for the $d_h$-dimensional attention-head.  The output of the cross-attention mechanism is computed as:
\begin{align}
\br_i =\sum_{j \in \mC} \alpha_{ij} \cdot \mathbf{W}_v [\bx_j \| \by_j]\,, \label{eq:attn_output} \forall i \in \mT,
\end{align}
where $\|$ denotes the concatenation of two vectors.
$\mathbf{W}_v \in \real^{d_h \times (d_x+d_y)}$ is a learnable weight matrix and $d_y$ is the dimension of the labels. In a $C$-class classification problem, we use the one-hot representation of the labels, so $d_y=C$.
Note that, though we present one-layer cross-attention with one attention-head for notational simplicity here, as shown in~\cite{vaswani2017attention}, this module can be generalized by considering  multi-head attention and by stacking multiple attention layers for larger capacity. 

The output of the attention mechanism is stacked with the target representation to form $\tilde{\bx}_i = [\bx_i \| \br_i]$.
In a classification problem, we model
\begin{align}
q_{\phi}^{emb}(\by_i = c|\bx_i,D^{emb}_{\mC})= \text{Softmax}_c \Bigg(h_{\psi}(\tilde{\bx}_i) \Bigg)\,,\label{eq:ppd_classification}\,
\end{align}
where $h_{\psi}:\real^{d_x+d_h} \rightarrow \real^{d_y}$ is a \emph{multi-layer perceptron} (MLP). In a regression setting, we use two separate MLPs $\mu_{\psi}$ and $\Sigma_{\psi'}$ to model the mean and covariance of the PPD, respectively,
\begin{align}
q_{\phi}^{emb}(\by_i|\bx_i,D^{emb}_{\mC})= \mathcal{N}\Bigg(\by_i ;\mu_{\psi}(\tilde{\bx}_i), \Sigma_{\psi'}(\tilde{\bx}_i) \Bigg)\,.\label{eq:ppd_regression}
\end{align}

\subsection{Loss Function}
We consider the minimization of the expected KL divergence (or equivalently cross entropy $H(\cdot, \cdot)$) between the true and the approximate PPD, where the expectation is w.r.t. the distribution of the  context data. The loss function is defined as:
\begin{align}
&\ell(\lambda; \mD_{\mL}) \nonumber\\ &{=}\mathbb{E}_{\mC}\Big[\frac{1}{|\mL|}\sum_{i \in \mL}H\Big(p_{true}(\by_i|\mG_i, D_{\mC}^{graph}),q_{\lambda}(\by_i|\mG_i,D_{\mC}^{graph})\Big)\Big]\,,\nonumber\\
&{\approx} \frac{-1}{|\mT|}\sum_{i \in \mT}\log q_{\lambda}(\by_i|\mG_i, D_{\mC}^{graph})\,,\label{eq:loss}
\end{align}
where the stochastic approximation of the loss in eq.~\eqref{eq:loss} is formed by random sampling of $\mT, \mC \subset \mL$ and computing the negative logarithm of the PPD using a forward pass through the model. This approximation is minimized to learn $\lambda$ using SGD via backpropagation.

\subsection{Inference}
Once the model parameters $\widehat{\lambda}$ are learned, we sample $P$ different context sets $\mC_{1:P} \subset \mL$ from the training data and form the prediction for a test set graph $\mG_i$ by computing the Monte Carlo average of the PPD to account for the uncertainty in context sampling as follows:
\begin{align}
q_{\widehat{\lambda}}(\by_i|\mG_i) &= \mathbb{E}_{\mC}\Bigg[ q_{\widehat{\lambda}}(\by_i|\mG_i, D_{\mC}^{graph}) \Bigg]\,,\nonumber\\
&\approx \frac{1}{P} \sum_{p=1}^P q_{\widehat{\lambda}}(\by_i|\mG_i, D_{\mC_p}^{graph})\,.\label{eq:pfn_pred}
\end{align}

The overall procedure for model training and inference is summarized in Algorithm~\ref{alg:train_test}. A graphical illustration is provided in Figure~\ref{fig:graphppd}.
\begin{algorithm}[h!]
\caption{Model training and inference}
\label{alg:train_test}
\begin{algorithmic}[1]
\STATE {\bfseries Input:} Training and test data: $\mD_{\mL}=\{\mG_i, \by_i\}_{i \in \mL}$, $\{\mG_i\}_{i \in \overline{\mL}}$.
\STATE {\bfseries Architecture:}  A feature extractor $g_{\theta}(\cdot)$, an approximate PPD module $q_{\phi}^{emb}$.
\STATE{\bfseries Hyperparameters:} Number of training iterations $N_{iter}$, learning rate sequence $\{\epsilon_j\}_{j=1}^{N_{iter}}$.
\STATE {\bfseries Output:}  Predictions on test set: $\{\widehat{q}_{\widehat{\lambda}}(\by_i|\mG_i)\}_{i \in \overline{\mL}}$ 
\STATE {\bfseries Initialization:}  Initialize $\theta=\theta^{(0)}$ and $\phi=\phi^{(0)}$ randomly. Form $\lambda^{(0)}=\{\theta^{(0)}, \phi^{(0)}\}$.
\STATE {\bfseries Model training:} Set $j=1$.
\WHILE{$j \leqslant N_{iter}$}
\STATE Sample target set indices $\mT \subset \mL$ and context set indices $\mC \subset \mL$ randomly.  
\STATE Compute $\bx_i=g_{\theta^{(j-1)}}(\mG_i)$, for all $i \in \mT \cup \mC$.
\STATE Compute $q_{\lambda^{(j-1)}}(\by_i|\mG_i, D_{\mC}^{graph})$ from eq.~\eqref{eq:ppd_graph} for all $i \in \mT$ using a forward pass through $q_{\phi^{(j-1)}}^{emb}$.   
\STATE Compute the gradient of the stochastic approximation of loss $\tilde{\ell}(\lambda;\mT, \mC) {=}\frac{-1}{|\mT|}\sum_{i \in \mT}\log q_{\lambda}(\by_i|\mG_i, D_{\mC}^{graph})$ w.r.t. $\lambda$ at $\lambda=\lambda^{(j-1)}$.
\STATE Update the model parameters $\lambda$ using SGD algorithm: $\lambda^{(j)} = \lambda^{(j-1)} -\epsilon_j \nabla_{\lambda} \tilde{\ell}(\lambda;\mT, \mC)\big|_{\lambda=\lambda^{(j-1)}}$.  
\ENDWHILE
\STATE Save the estimated model $\widehat{\lambda}= \lambda^{(N_{iter})}$.
\STATE For each $i \in  \overline{\mL}$, sample $P$ different sets of random context indices independently, $\mC_{1:P} \subset  \mL$, and compute probabilistic prediction $\widehat{q}_{\widehat{\lambda}}(\by_i|\mG_i) = \frac{1}{P} \sum_{p=1}^P q_{\widehat{\lambda}}(\by_i|\mG_i, D_{\mC_p}^{graph})$.  
\end{algorithmic}
\end{algorithm}

\subsection{Discussion}
The PPD-learning framework presented in this section is general in many regards. For example, existing graph-level representation learning models using GNNs~\cite{xu2019,baek2021} and graph transformers~\cite{kreuzer2021rethinking,rampavsek2022recipe} can be directly incorporated as the graph encoder $g_{\theta}$ inside our framework. Similarly, various architectures ranging from prototypical networks~\cite{snell2017proto}, conditional neural process variants~\cite{garnelo2018b,nguyen2022tnp,lee2020bnp}, to transformers without positional encoding~\cite{muller2021} can be used for modelling $q_{\phi}^{emb}$. The overall framework is applicable to other domains as well if a suitable feature extractor is used. 


The pseudocode in Algorithm~\ref{alg:train_test} describes \emph{end-to-end} (E2E) training for joint optimization of the graph encoder parameter $\theta$ and the PPD parameters $\phi$. However, if a pre-trained encoder is available, we could freeze its parameters and optimize $\phi$ only. In that case, if we precompute and save the embedding for each graph in the dataset, we can benefit from reduced training time, since the requirement of a potentially expensive forward pass through a complex graph encoder is alleviated.

\begin{table*}[htbp]
\centering
\caption{Mean and standard error of classification ROC-AUC ($\uparrow$, in \%) for 7 benchmark OGB datasets. For each feature extractor, the better result is shown in bold.}
\label{tab:results_ogb_main}
\resizebox{\columnwidth*2}{!}{
\begin{tabular}{l|ccccccc|c|c}
\toprule
Alg. &BACE         &BBBP     &CLINTOX        &HIV &SIDER &TOXCAST  &TOX21 &\#(win/tie/loss) &rel. improvement   \\ \midrule
GINE   &73.5$\pm$2.0 &68.0$\pm$1.6 &86.0$\pm$2.4 &\textbf{75.7$\pm$2.0} &55.3$\pm$1.9 &60.9$\pm$0.5 &73.2$\pm$0.9  &- &-\\ 
Ours   &\textbf{74.2$\pm$3.2} &\textbf{69.8$\pm$2.1} &\textbf{86.7$\pm$2.4} &75.0$\pm$1.3 &\textbf{57.8$\pm$1.1}* &\textbf{61.9$\pm$0.7}* &\textbf{74.1$\pm$0.8}  &6/0/1 &1.59\%\\ \hline
GMT  &67.9$\pm$6.8 &67.3$\pm$0.8 &82.3$\pm$5.5 &\textbf{76.1$\pm$2.1} &53.4$\pm$3.1 &63.8$\pm$0.6 &75.7$\pm$0.6 &- &- \\
Ours  &\textbf{76.6$\pm$5.4}* &\textbf{68.1$\pm$1.4} &\textbf{85.5$\pm$2.9} &75.5$\pm$1.0 &\textbf{56.8$\pm$1.8}* &\textbf{66.2$\pm$0.7}* &\textbf{75.8$\pm$0.4} &6/0/1 &3.93\%\\
\bottomrule
\end{tabular}}
\end{table*}

\begin{table*}[htbp]
\centering
\caption{Mean and standard error of classification accuracy ($\uparrow$, in \%) for 10 benchmark TU datasets. For each feature extractor, the better result is shown in bold.}
\label{tab:results_tu_main}
\resizebox{\columnwidth*2}{!}{
\begin{tabular}{l|ccccc|ccccc|c|c}
\toprule
Alg. &D\&D         &PROTEINS     &MUTAG        &NCI1       &PTC       &IMDB-B     &IMDB-M        &REDDIT-B     &REDDIT-M       &COLLAB  &\#(win/tie/loss) &rel. improvement  \\ \midrule
GIN    &72.1$\pm$1.5 &72.0$\pm$1.2 &83.4$\pm$1.5 &78.3$\pm$0.5 &54.6$\pm$1.4 &74.3$\pm$0.5 &51.2$\pm$0.4 &91.5$\pm$0.4 &55.6$\pm$0.6 &81.1$\pm$0.4 &- &-\\ 
Ours
&\textbf{72.2$\pm$1.4} &72.0$\pm$1.3 &\textbf{84.0$\pm$3.3} &\textbf{78.7$\pm$0.4}* &\textbf{54.7$\pm$1.6} &74.3$\pm$0.7 &51.2$\pm$0.6 &\textbf{91.8$\pm$0.8} &\textbf{56.5$\pm$0.3}* &\textbf{81.7$\pm$0.4}*  &7/3/0 &0.44\%\\ 
\midrule
GMT    &78.3$\pm$0.5 
&74.8$\pm$0.9
&\textbf{82.7$\pm$0.6} 
&76.3$\pm$0.4 
&56.0$\pm$2.7 &\textbf{73.7$\pm$0.8} &\textbf{50.6$\pm$0.5} &91.9$\pm$0.2 &55.7$\pm$0.3 &80.4$\pm$0.3 &- &-\\
Ours
&\textbf{78.4$\pm$0.4} &\textbf{74.9$\pm$0.6} &81.3$\pm$1.1 &\textbf{76.6$\pm$0.4} &\textbf{57.2$\pm$1.7} &73.5$\pm$0.8 &50.1$\pm$0.9 &91.9$\pm$0.6 
&\textbf{56.1$\pm$0.5} &\textbf{81.3$\pm$0.3}*  &6/1/3  &0.19\%\\
\bottomrule
\end{tabular}}
\end{table*}
\begin{table}[htbp]
\centering
\caption{Mean and standard error of MAE ($\downarrow$) for ZINC-12k dataset. For each feature extractor, the better result is shown in bold.}
\label{tab:results_zinc_main}
\footnotesize
\setlength{\tabcolsep}{2pt}
\begin{tabular}{cc|cc}
\toprule
GINE &Ours &GraphGPS &Ours \\ \midrule
0.070$\pm$0.004  &\textbf{0.065$\pm$0.004}* &0.070$\pm$0.004 &\textbf{0.067$\pm$0.003}* \\ \hline
rel. improvement  &7.7\% & - &4.5\% \\ 
\bottomrule
\end{tabular} 
\end{table}

\section{Relationship to Prior Work}
\label{sec:related_work}
 Our work is related to i) uncertainty characterization for graph neural networks, ii) graph-level representation learning using GNNs and graph transformers, and iii) conditional predictive density estimation using neural architectures. 
 

Over the last few years, there has been significant research exploring the generation of uncertainty-aware predictions from Graph Neural Networks (GNNs)~\cite{defferrard2016,kipf2017} for node-level tasks (e.g. node classification~\cite{ng2018,zhang2019,ma2019,pal2020,liu2020,hasanzadeh2020}), edge-level tasks (e.g. link prediction~\cite{pal2020,opolka2022}, and recommendation system tasks~\cite{sun2020}). 
These approaches rely on statistical modelling~\cite{zhang2019}, variational inference~\cite{ma2019,elinas2020}, and \emph{maximum a posteriori} (MAP) estimation~\cite{pal2020} of the unobserved `true' graph topology conditioned on the observed graph and other available data such as node features and training labels.
Other works consider Gaussian process based designs~\cite{ng2018,opolka2022}, modelling of GNNs' weight uncertainty~\cite{hasanzadeh2020}, and latent variable modelling for introducing stochasticity in GNNs~\cite{liu2020}. However, all of these approaches are applied to transductive node classification or link prediction tasks, whereas the graph-level tasks we address here are fundamentally inductive in nature. So, none of these approaches are suitable for direct application in our problem setting. Moreover, in node classification and link prediction, suitable modelling of the uncertainty associated with the graph structure can provide useful inductive bias, whereas in graph classification tasks, it is not so obvious.

In recent years, there has been extensive focus on learning effective  representations for solving graph-level classification and regression.
 In these architectures, either GNNs~\cite{xu2019,baek2021,dwivedi2021graph,wu2022,duval2022} or graph transformers~\cite{ying2021, kreuzer2021rethinking, rampavsek2022recipe} are employed to learn node embeddings, which are successively fed to a pooling layer to obtain a single graph embedding. 
These models primarily differ from each other in their design of these two components. 
However, our contribution is complementary in nature, since the architecture design of deterministic GNNs or graph transformers is not our focus. Our PPD learning approach is flexible in the sense that the existing graph representation learning architectures can be used directly in our framework. 

In our work, we have another neural component, which learns to approximate the PPD using a single forward pass.
Similar models have been proposed in meta-learning~\cite{gordon2018metalearning,muller2021}, few-shot learning~\cite{snell2017proto}, and neural processes variants~\cite{garnelo2018,kim2019,lee2020bnp,lee2020rnp,nguyen2022tnp}. In these cases, 
using a training set of relatively small scale datasets, one strives to learn a prediction model, which would have generalization capability to unseen, related datasets. Instead of Bayesian inference of the model parameters, such models aim to approximate the PPD directly by adopting suitable architectural designs, such that, aside from the small set of instances whose labels are to be estimated (termed queries in few-shot learning and target samples in the neural process literature), the labeled dataset (called support samples in few-shot learning and context data in neural process terminology) associated with them can also serve as inputs to the model. Besides the difference in architecture design and application, one fundamental difference of our approach from this body of work is that our technique uses target and context data sampling from a single training dataset for PPD modelling and is employed in a traditional supervised setting. Our work is more generally applicable since the probabilistic framework we propose can be easily extended to other domains by a suitable design of the instance level embedding method. 

Recently, there has been some work that combines neural processes with GNNs~\cite{carr2019,day2020}. However, these approaches are considerably different from our work since they consider a transductive setting for node classification and edge-imputation.   



\begin{table*}[ht]
\centering
\caption{Mean and standard error of classification ROC-AUC ($\uparrow$, in \%) of MC Dropout and our approach for 7 benchmark OGB datasets. For each feature extractor, the better result is shown in bold.}
\label{tab:results_ogb_mc}
\scriptsize
\setlength{\tabcolsep}{2pt}
\begin{tabular}{l|ccccccc|c|c}
\toprule
Alg. &BACE         &BBBP     &CLINTOX        &HIV &SIDER &TOXCAST  &TOX21 &\#(win/tie/loss) &rel. improvement   \\ \midrule
GINE-MC &73.4$\pm$2.1 &67.9$\pm$1.6 &86.3$\pm$2.7 &\textbf{75.6$\pm$2.0} &55.3$\pm$1.7 &60.8$\pm$0.5 &73.1$\pm$1.0    &- &-\\ 
Ours   &\textbf{74.2$\pm$3.2} &\textbf{69.8$\pm$2.1}* &\textbf{86.7$\pm$2.4} &75.0$\pm$1.3 &\textbf{57.8$\pm$1.1}* &\textbf{61.9$\pm$0.7}* &\textbf{74.1$\pm$0.8}  &6/0/1 &1.61\%\\ \hline
GMT-MC  &67.9$\pm$6.7 &67.3$\pm$0.8 &83.2$\pm$4.4 &\textbf{76.2$\pm$2.1} &53.4$\pm$2.9 &63.8$\pm$0.6 &\textbf{75.9$\pm$0.6}  &- &- \\
Ours  &\textbf{76.6$\pm$5.4}* &\textbf{68.1$\pm$1.4} &\textbf{85.5$\pm$2.9} &75.5$\pm$1.0 &\textbf{56.8$\pm$1.8}* &\textbf{66.2$\pm$0.7}* &75.8$\pm$0.4 &5/0/2 &3.72\%\\
\bottomrule
\end{tabular}
\end{table*}

\begin{table*}[ht]
\centering
\caption{Mean and standard error of classification accuracy ($\uparrow$, in \%) of MC Dropout and our approach for 10 benchmark TU datasets. For each feature extractor, the better result is shown in bold.}
\label{tab:results_tu_mc}
\scriptsize
\setlength{\tabcolsep}{1.5pt}
\begin{tabular}{l|ccccc|ccccc|c|c}
\toprule
Alg. &D\&D         &PROTEINS     &MUTAG        &NCI1       &PTC       &IMDB-B     &IMDB-M        &REDDIT-B     &REDDIT-M       &COLLAB  &\#(win/tie/loss) &rel. improvement  \\ \midrule
GIN-MC &\textbf{72.3$\pm$1.5} &71.6$\pm$1.4 &83.4$\pm$1.7 &78.1$\pm$0.5 &54.5$\pm$1.3 &74.2$\pm$0.5 &51.2$\pm$0.5 &91.5$\pm$0.3 &55.5$\pm$0.6 &81.2$\pm$0.3  &- &-\\ 
Ours
&72.2$\pm$1.4 &\textbf{72.0$\pm$1.3} &\textbf{84.0$\pm$3.3} &\textbf{78.7$\pm$0.4}* &\textbf{54.7$\pm$1.6} &\textbf{74.3$\pm$0.7} &51.2$\pm$0.6 &\textbf{91.8$\pm$0.8} &\textbf{56.5$\pm$0.3}* &\textbf{81.7$\pm$0.4}*  &8/1/1 &0.54\%\\ 
\midrule
GMT-MC &78.3$\pm$0.5 &74.9$\pm$0.7 &\textbf{82.6$\pm$0.7}* &76.3$\pm$0.4 &56.2$\pm$3.0 &\textbf{73.8$\pm$0.7} &\textbf{50.5$\pm$0.7} &91.9$\pm$0.2 &55.7$\pm$0.3 &80.5$\pm$0.3     &- &-\\
Ours
&\textbf{78.4$\pm$0.4} &74.9$\pm$0.6 &81.3$\pm$1.1 &\textbf{76.6$\pm$0.4} &\textbf{57.2$\pm$1.7} &73.5$\pm$0.8 &50.1$\pm$0.9 &91.9$\pm$0.6 
&\textbf{56.1$\pm$0.5} &\textbf{81.3$\pm$0.3}*  &5/2/3  &0.13\%\\
\bottomrule
\end{tabular}
\end{table*}

\begin{table*}[ht]
\centering
\caption{Mean and standard error of classification ROC-AUC ($\uparrow$, in \%) of Ensemble and our approach for 7 benchmark OGB datasets. For each feature extractor, the better result is shown in bold.}
\label{tab:results_ogb_ensmb}
\scriptsize
\setlength{\tabcolsep}{2pt}
\begin{tabular}{l|ccccccc|c|c}
\toprule
Alg. &BACE         &BBBP     &CLINTOX        &HIV &SIDER &TOXCAST  &TOX21 &\#(win/tie/loss) &rel. improvement   \\ \midrule
GINE-Ensmb. &74.4$\pm$1.8 &69.0$\pm$1.1 &88.3$\pm$1.8 &\textbf{78.0$\pm$0.7}* &56.8$\pm$1.5 &61.8$\pm$0.5 &73.8$\pm$0.4
    &- &-\\ 
Ours   &74.2$\pm$3.2 &69.8$\pm$2.1 &86.7$\pm$2.4 &75.0$\pm$1.3 &57.8$\pm$1.1* &61.9$\pm$0.7 &74.1$\pm$0.8  &4/0/3 &-0.31\% \\ \hline
Ours-Ensmb.  &\textbf{75.9$\pm$2.3} &\textbf{72.3$\pm$2.1} &\textbf{88.4$\pm$1.2} &76.4$\pm$1.0 &\textbf{58.8$\pm$0.8} &\textbf{62.9$\pm$0.5} &\textbf{75.6$\pm$0.6} &6/0/1  &1.82\%  \\ \hline
GMT-Ensmb. &68.7$\pm$4.5 &68.2$\pm$1.4 &86.2$\pm$3.5 &\textbf{77.1$\pm$1.0}* &55.1$\pm$2.5 &64.3$\pm$0.6 &76.6$\pm$0.2*
  &- &- \\ 
Ours  &76.6$\pm$5.4* &68.1$\pm$1.4 &85.5$\pm$2.9 &75.5$\pm$1.0 &56.8$\pm$1.8 &66.2$\pm$0.7* &75.8$\pm$0.4 &3/0/4 &1.95\% \\ \hline
Ours-Ensmb   &\textbf{79.4$\pm$2.2} &\textbf{69.4$\pm$1.0} &\textbf{87.3$\pm$2.1} &76.1$\pm$0.7 &\textbf{57.4$\pm$1.3} &\textbf{67.3$\pm$0.3} &\textbf{76.7$\pm$0.4} &6/0/1  &3.75\% \\
\bottomrule
\end{tabular}
\end{table*}

\begin{table*}[htbp]
\centering
\caption{Mean and standard error of classification accuracy ($\uparrow$, in \%) of Ensemble and our approach for 10 benchmark TU datasets. For each feature extractor, the better result is shown in bold.}
\label{tab:results_tu_ensmb}
\scriptsize
\setlength{\tabcolsep}{1.5pt}
\begin{tabular}{l|ccccc|ccccc|c|c}
\toprule
Alg. &D\&D         &PROTEINS     &MUTAG        &NCI1       &PTC       &IMDB-B     &IMDB-M        &REDDIT-B     &REDDIT-M       &COLLAB  &\#(win/tie/loss) &rel. improvement  \\ \midrule
GIN-Ensmb.  &75.0$\pm$0.7 &73.1$\pm$0.8 &83.5$\pm$1.1 &80.3$\pm$0.4 &55.0$\pm$1.2 &74.1$\pm$0.6 &51.7$\pm$0.4 &91.6$\pm$0.2 &56.1$\pm$0.3 &81.6$\pm$0.2    &- &-\\ 
Ours
&72.2$\pm$1.4 &72.0$\pm$1.3 &84.0$\pm$3.3 &78.7$\pm$0.4 &54.7$\pm$1.6 &74.3$\pm$0.7 &51.2$\pm$0.6 &91.8$\pm$0.8 &56.5$\pm$0.3 &81.7$\pm$0.4  &5/0/5 & -0.67\%
 \\ \hline
Ours-Ensmb.  &\textbf{77.2$\pm$0.8} &\textbf{73.7$\pm$0.9} &\textbf{86.6$\pm$0.8} &\textbf{80.8$\pm$0.3} &\textbf{55.5$\pm$1.9} &\textbf{74.4$\pm$0.5} &51.7$\pm$0.4 &\textbf{92.2$\pm$0.2} &\textbf{57.0$\pm$0.3} &\textbf{82.5$\pm$0.2} &9/1/0 &1.27\% \\ 
\hline
GMT-Ensmb.  &\textbf{78.8$\pm$0.2} &75.5$\pm$0.6 &\textbf{82.8$\pm$0.8} &76.8$\pm$0.2 &56.6$\pm$1.3 &\textbf{74.1$\pm$0.6} &\textbf{51.2$\pm$0.3} &92.0$\pm$0.2 &55.9$\pm$0.2 &81.3$\pm$0.1   &- &-\\
Ours
&78.4$\pm$0.4 &74.9$\pm$0.6 &81.3$\pm$1.1 &76.6$\pm$0.4 &\textbf{57.2$\pm$1.7} &73.5$\pm$0.8 &50.1$\pm$0.9 &91.9$\pm$0.6 
&56.1$\pm$0.5 &81.3$\pm$0.3  &2/1/7  &-0.48\% \\ \hline
Ours-Ensmb. 
&78.7$\pm$0.6 &75.5$\pm$0.2 &81.6$\pm$1.1 &\textbf{77.5$\pm$0.2} &56.7$\pm$1.0 &74.0$\pm$0.7 &50.9$\pm$0.5 &\textbf{92.2$\pm$0.2} &\textbf{56.5$\pm$0.2} &\textbf{82.3$\pm$0.2} &5/1/4 &0.13\% \\
\bottomrule
\end{tabular}
\end{table*}

\begin{figure*}[h!]
\begin{minipage}{0.325\textwidth}
  \includegraphics[width=\linewidth]{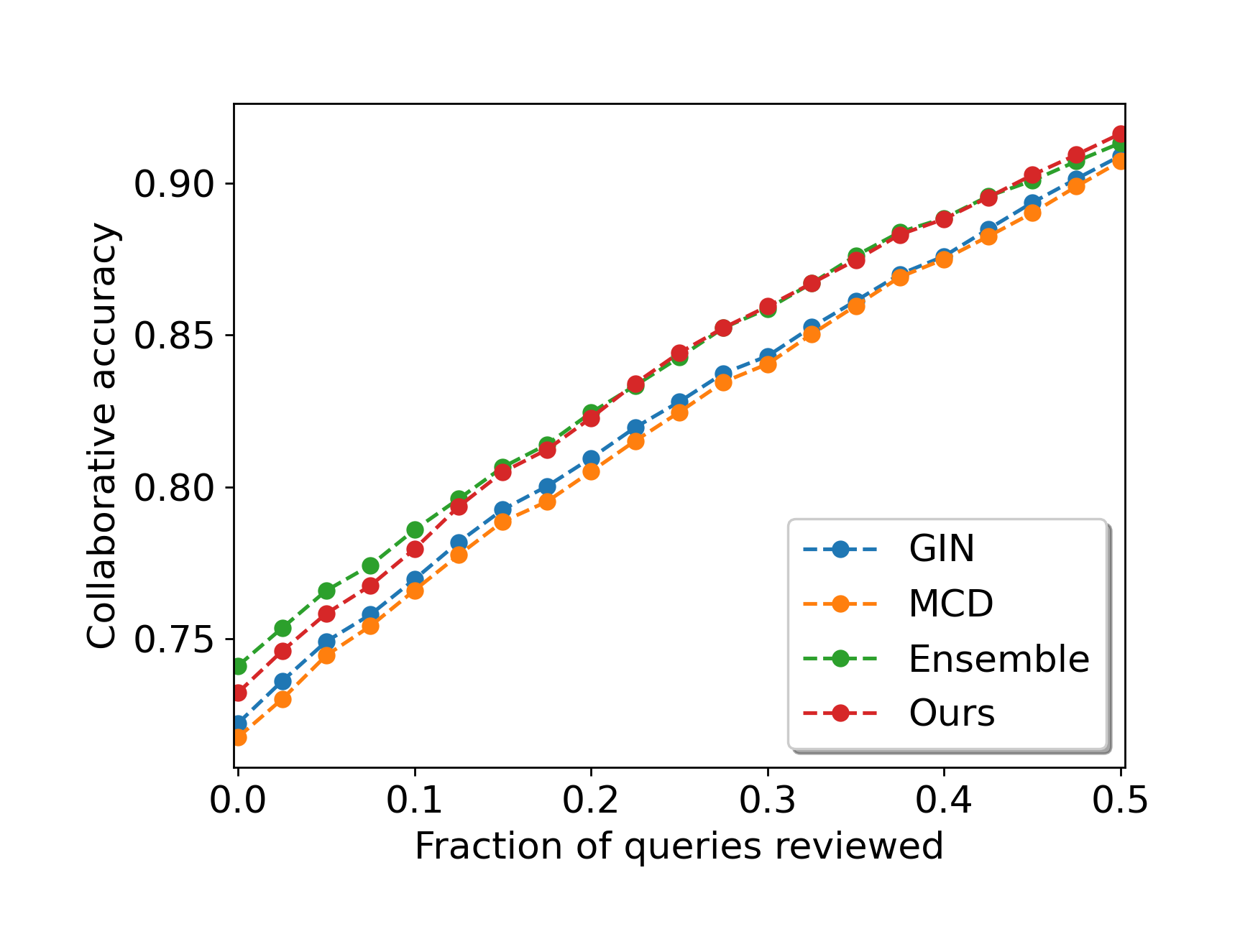}
  \subcaption{}
\end{minipage}%
\begin{minipage}{0.325\textwidth}
  \includegraphics[width=\linewidth]{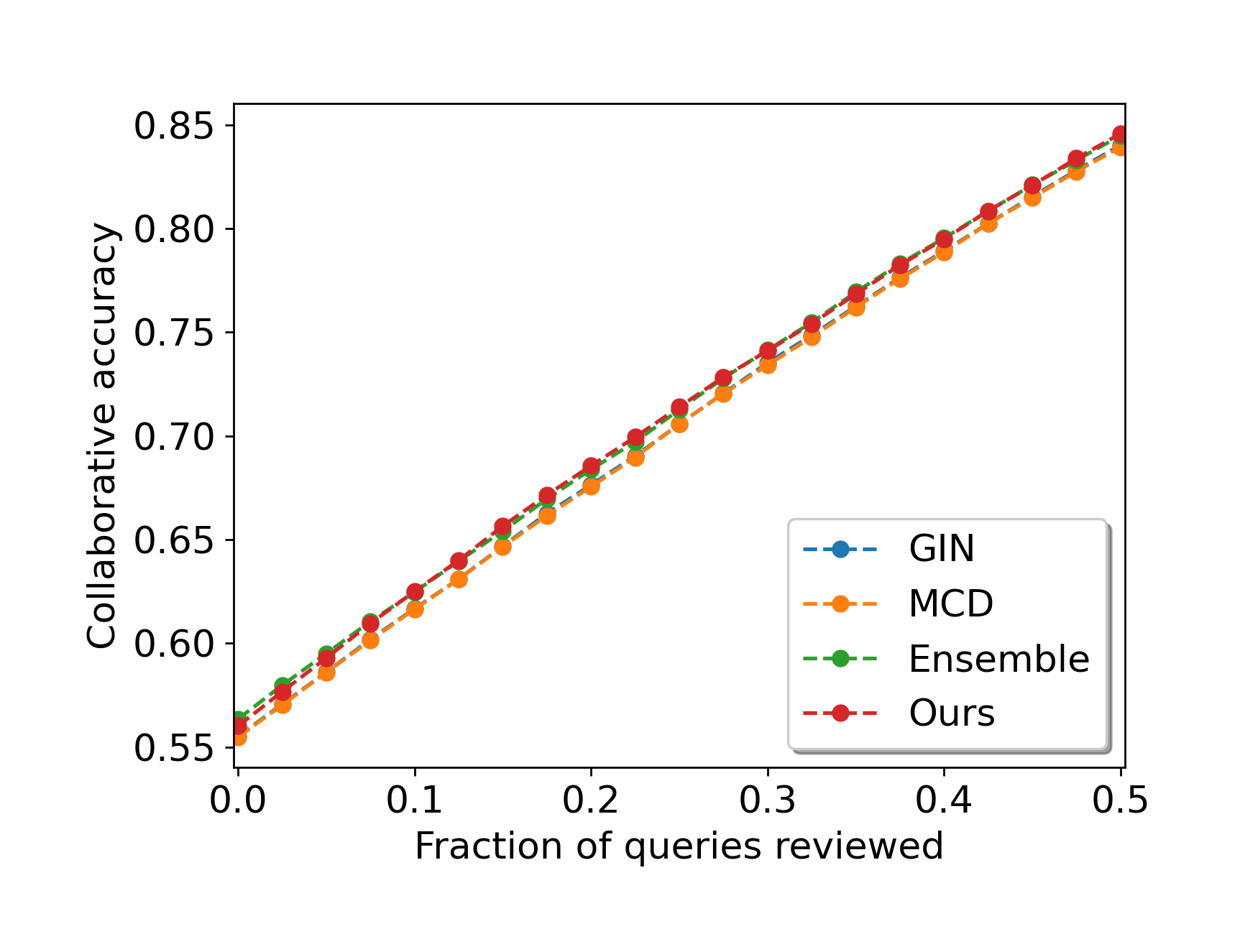}
  \subcaption{}
\end{minipage}  
\begin{minipage}{0.325\textwidth}
  \includegraphics[width=\linewidth]{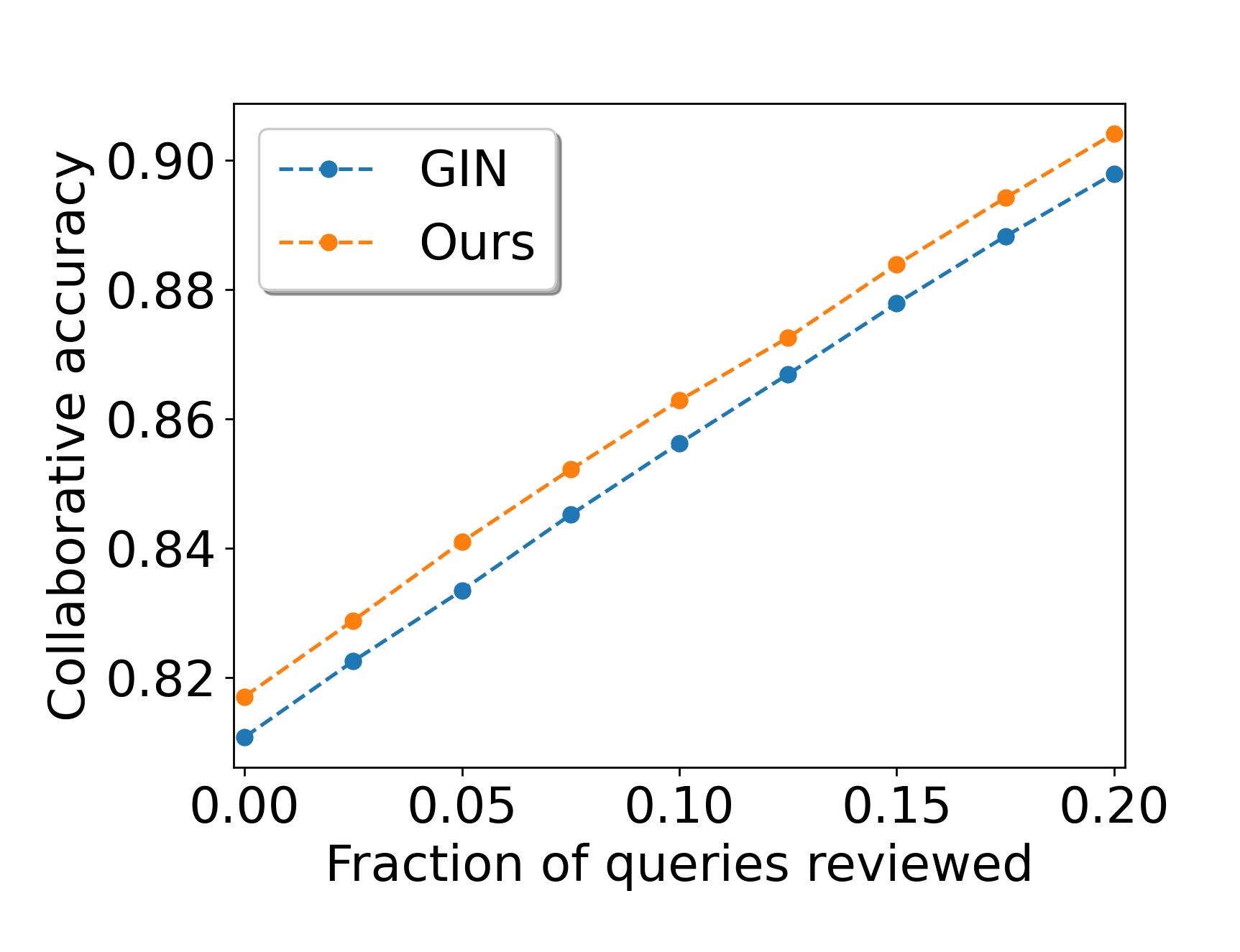}
  \subcaption{}
\end{minipage} %
\begin{minipage}{0.325\textwidth}
  \includegraphics[width=\linewidth]{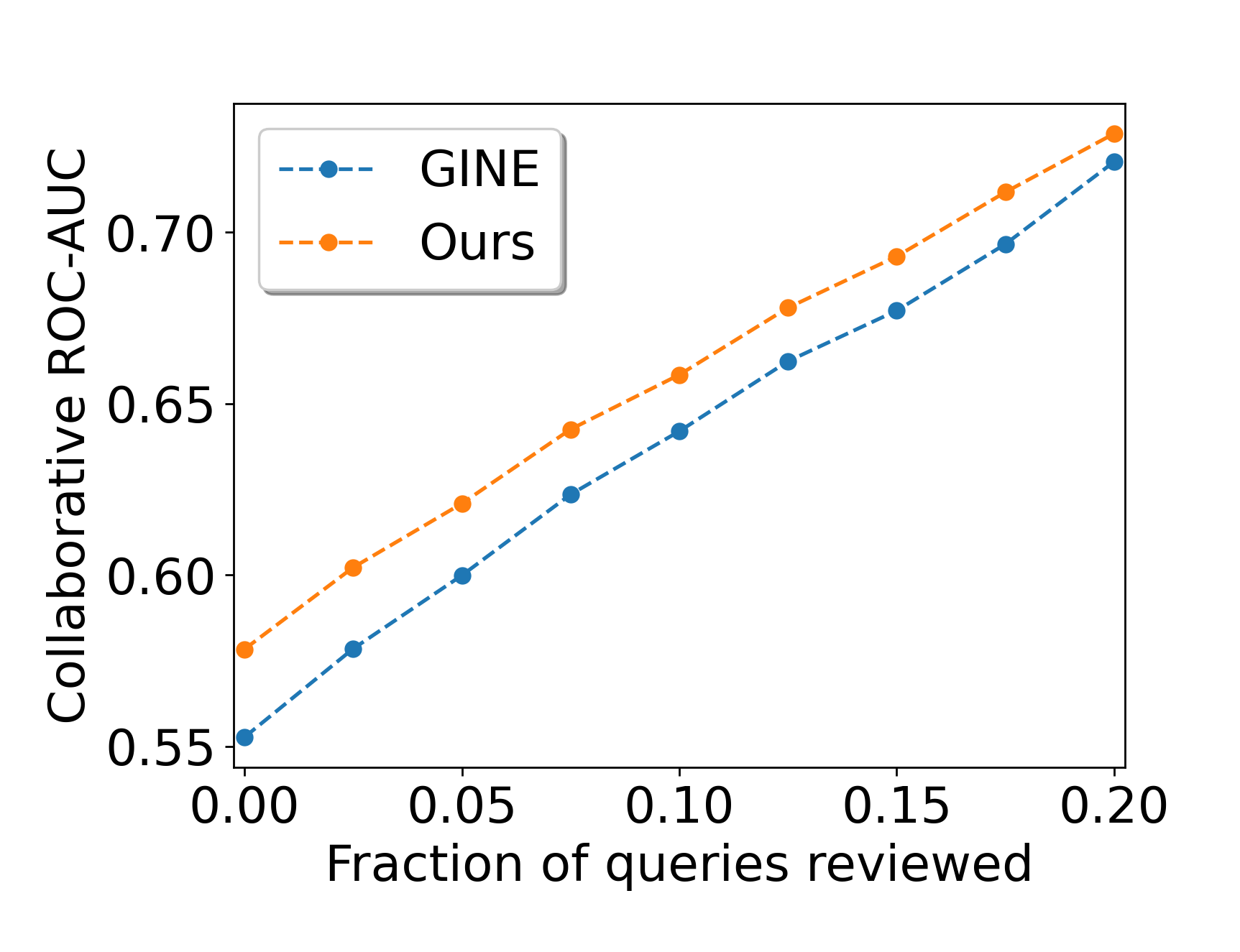}
  \subcaption{}
\end{minipage}%
\begin{minipage}{0.325\textwidth}
  \includegraphics[width=\linewidth]{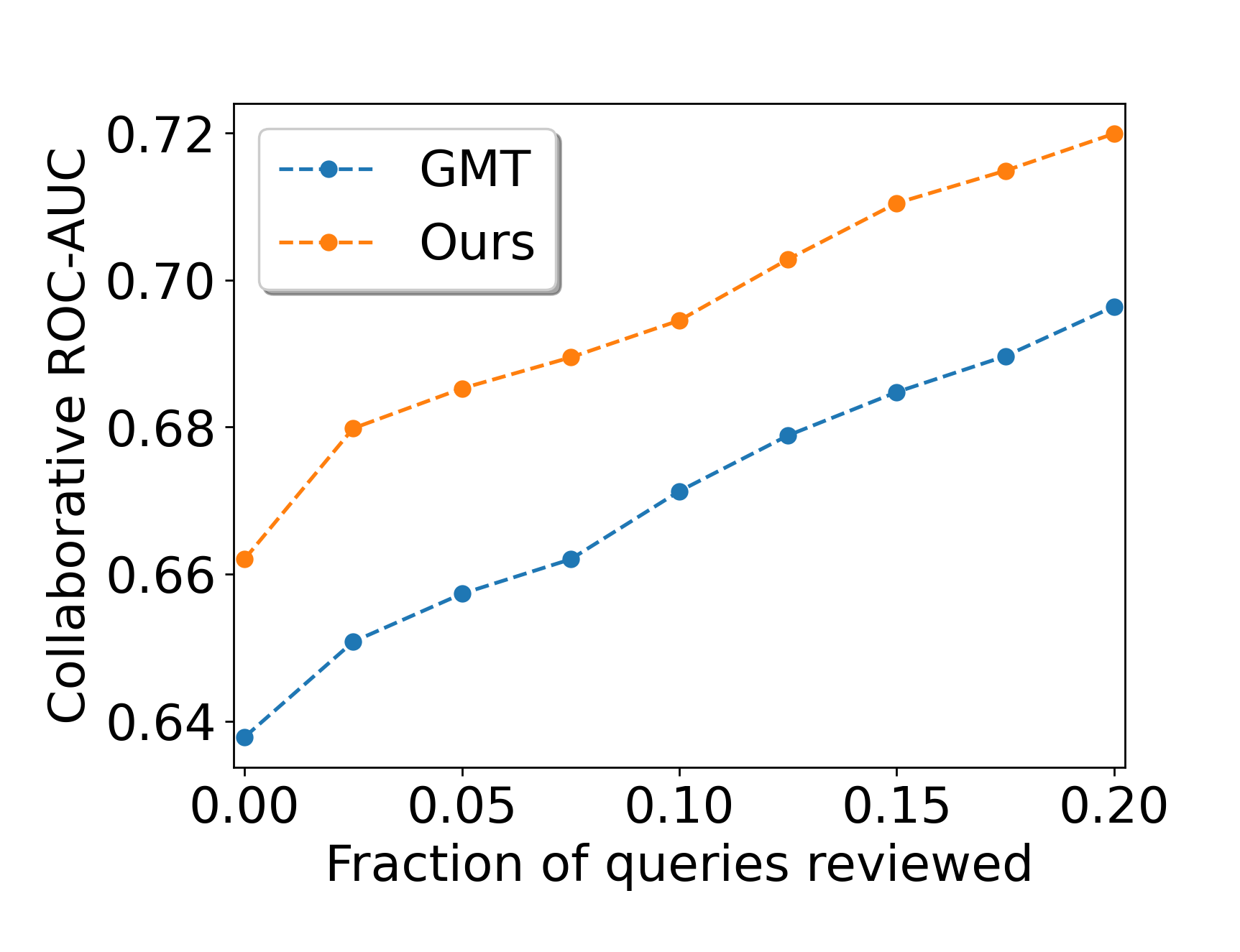}
  \subcaption{}
\end{minipage}  
\begin{minipage}{0.325\textwidth}
  \includegraphics[width=\linewidth]{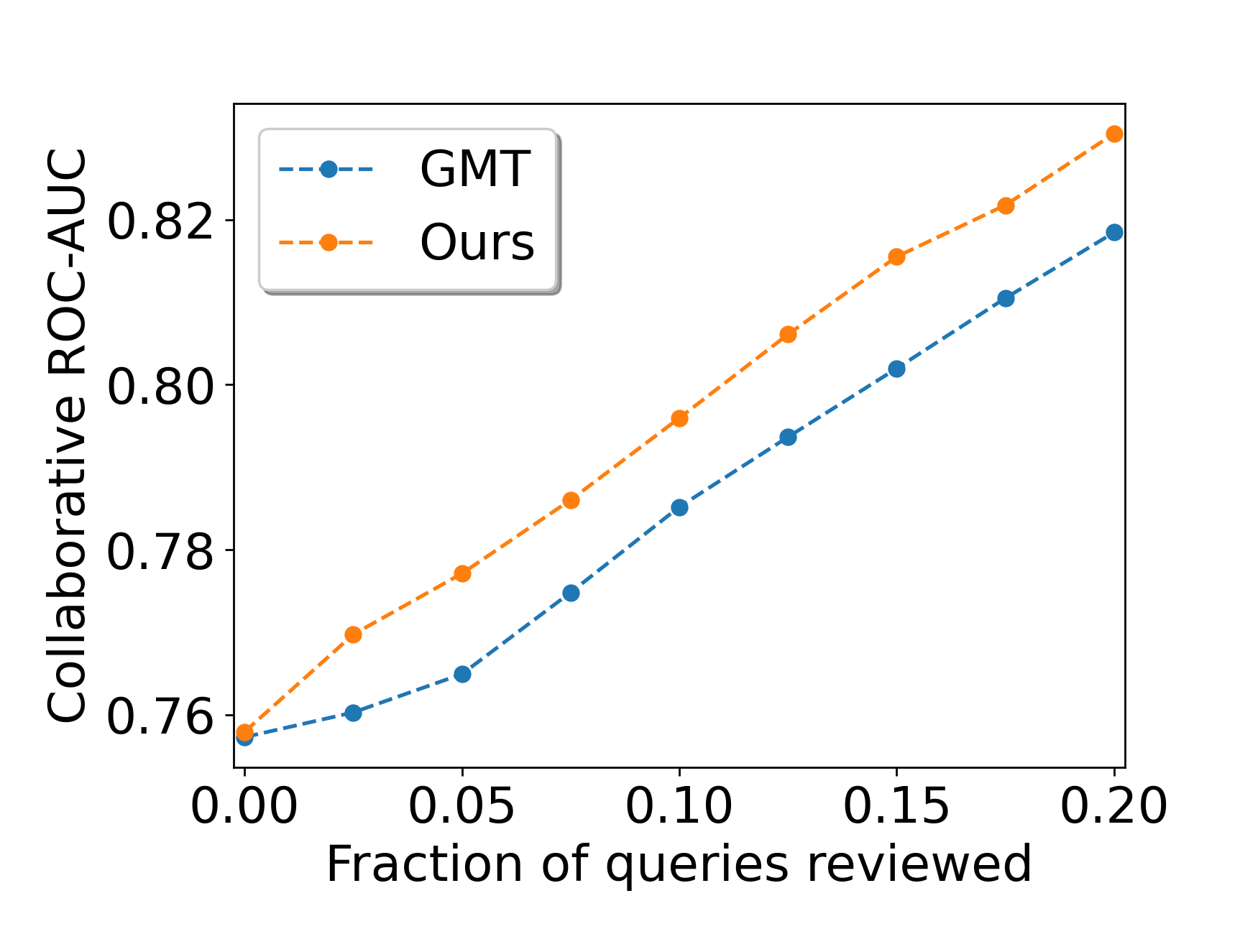}
  \subcaption{}
\end{minipage} %
 \caption{Selective prediction results on a) PROTEINS, b) REDDIT-M, c) COLLAB, d) SIDER, e) TOXCAST, and f) TOX21 datasets.}
 \label{fig:sel_pred}
\end{figure*}

\section{Experimental Results}
\label{sec:results}
In order to demonstrate the effectiveness of the proposed framework, we perform graph classification experiments on several benchmarks including 7 OGB datasets~\cite{hu2021OpenGraphBenchmark} and 10 TU datasets~\cite{morris2020tudataset}. 
Additionally, we tackle a graph regression benchmark on ZINC-12k~\cite{dwivedi2020benchmarkgnns}.  

\subsection{Datasets and Evaluation Metrics}
 We use 7 molecular datasets from the OGB benchmarks, namely, BACE, BBBP, CLINTOX, HIV, SIDER, TOXCAST, and TOX21. The prediction performance is measured by ROC-AUC for these datasets. Additionally, we choose 10 TU datasets including 
5 bioinformatics datasets (D\&D, PROTEINS, MUTAG, NCI1, and PTC-MR), and 5 social network datasets (COLLAB, IMDB-BINARY, IMDB-MULTI, REDDIT-BINARY, REDDIT-MULTI-5k, and COLLAB). 
For these datasets, the standard evaluation metric is classification accuracy. 
The regression task on ZINC-12k is evaluated using \emph{mean absolute error} (MAE). The detailed overview and statistics of these datasets is summarized in Table.~\ref{tab:dataset_stat} in Appendix~\ref{sec:app:dataset}.

\subsection{Training Setup and Hyperparameters}

\textbf{Data splitting:} For the OGB benchmarks and ZINC-12k datasets, we use the publicly available splits to run 10 times with different random seeds for the initialization on each model in different trials. 
We conduct 10-fold cross-validation for 10 trials on the TU datasets and report the average accuracy along with its standard error.

\textbf{Graph encoders:} We use three different graph encoders to highlight the general applicability of our GraphPPD framework. First, we use the Graph Isomorphism Network (GIN)~\cite{xu2019}, which is one of the most widely used GNNs on graph-level tasks and a more sophisticated Graph Multiset Transformer~\cite{baek2021} to instantiate the feature extractor module in our framework for the TU datasets. 
For the OGB and ZINC-12K datasets, we use a GIN variant with edge-features, called GINE~\cite{hu2020StrategiesPretrainingGraph}, since the utilization of edge features for learning is beneficial for these datasets.
We skip comparison to other graph classification baselines, such as~\cite{zhang2018,ying2018diffpool,lee2019,gao2019,wang2020haar,ranjan2020,bianchi2020mincutpool,morris2020tudataset,hu2021OpenGraphBenchmark,wu2022,duval2022}, since GMT is shown to either outperform these approaches or achieve comparable performance on these TU and OGB datasets.
We also consider a recently proposed graph transformer, GraphGPS~\cite{rampavsek2022recipe}, for the regression on ZINC-12k, since it achieves the state-of-the-art performance for this task.

\textbf{Training scheme:} For our proposed approach, we consider two different variants. The \emph{end-to-end} (E2E) setting corresponds to Algorithm~\ref{alg:train_test}, where $\theta$ and $\phi$ are learned jointly.
For the \emph{2-stage} version, we freeze $\theta$ after the baseline is trained, and use the pre-trained feature extractor provided graph embeddings for learning $\phi$ in isolation. In order to ensure a fair comparison between the baseline algorithms and their incorporation in our framework as feature extractor modules, we employ the original configurations for their architectures from the original papers~\cite{baek2021,dwivedi2021graph,rampavsek2022recipe}, as well as other hyperparameters such as batch size, learning rate and its decay schedule, weight decay, number of epochs, patience for early stopping, and maximum gradient norm for clipping.
We only tune the hyperparameters of the amortized PPD module, using a 10-fold cross-validation for the TU datasets and using the fixed validation split for the OGB and ZINC-12k datasets. Hyperparameter tuning for the PPD module is carried out in E2E mode only and we subsequently reuse the same values for the 2-stage training. The tuned hyperparameters of the PPD module used in our experiments are listed in Table~\ref{tab:hparam} in Appendix~\ref{sec:app:hyperparam}. For a concise presentation, we only provide the results from the E2E  training in this section and defer the results for the 2-stage mode to Tables~\ref{tab:results_ogb_base},~\ref{tab:results_tu_base}, and~\ref{tab:results_zinc_base} in Appendix~\ref{sec:app:detailed_res}, which also contains the comparison of different models in terms of training time and number of learnable parameters.
We use the Wilcoxon signed-rank test to compare our approach to the corresponding feature extractor model and indicate a significantly better result at the 5\% level with an asterisk.

Tables~\ref{tab:results_ogb_main},~\ref{tab:results_tu_main}, and~\ref{tab:results_zinc_main} summarize the results on the OGB, TU, and ZINC-12k datasets, respectively. We observe that in most cases, the proposed approach outperforms the corresponding feature extractor. The relative improvement offered by our approach is higher for the OGB and ZINC-12k datasets, compared to the TU datasets. Among the 10 TU datasets, many are relatively small-scale (few graphs). It is more challenging to train the GNN and the PPD module together on the small datasets, and the 2-stage version works considerably better in most of these cases (shown in Appendix~\ref{sec:app:detailed_res}).  

\subsection{Comparison with Monte Carlo Dropout}
We compare our approach with Monte Carlo (MC) Dropout~\cite{gal2016}, which provides a variational approximation of the posterior of the model parameters. The results in Tables~\ref{tab:results_ogb_mc} and~\ref{tab:results_tu_mc} demonstrate that approximation of the PPD using MC dropout fails to improve classification performance of the graph encoder models, whereas the superiority of the proposed technique is evident in most cases. 


\subsection{Comparison with Ensembles}
 Deep ensembles~\cite{lakshminarayanan2017ensemble} offer considerably better performance than the constituent models in isolation in many scenarios but frequently suffer from a large computation burden. For comparison of the proposed methods with ensembles, we choose the ensemble size as the lowest integer such that the ensemble has strictly more learnable parameters compared to our approach. Tables~\ref{tab:results_ogb_ensmb} and~\ref{tab:results_tu_ensmb} show that our approach performs slightly worse than the ensembles (except for the GMT-Ensembles on OGB datasets). However, our approach has fewer learnable parameters (far fewer for the GMT architecture particularly) compared to the ensemble in each case and it requires significantly lower training time in most cases. For example, our approach using the GIN encoder requires 39.5\% lower training time and 14.5\% fewer trainable parameters on average for the TU datasets (detailed comparison for all cases are provided in Appendix~\ref{sec:app:detailed_res}. If we form an ensemble using the same number of our models, it outperforms the ensemble of the feature extractor models in most cases. 
 

\subsection{Summary of Obtained Results}
We also conduct Wilcoxon signed rank tests across datasets and graph encodes, which reveal that a) E2E and 2-stage versions of the proposed GraphPPD and Ensemble significantly outperforms the graph encoders and MC Dropout, b) the difference in performance between MC dropout and the graph encoders is not statistically significant, and, c) the difference in performance between any version of GraphPPD and Ensemble is not statistically significant. In each case, the significance is declared at the 5\% level.  

\subsection{Selective Prediction}
In this setting, the model evaluation protocol follows a collaborative strategy, which requires the presence of an oracle reviewer (moderator), who can be asked 
 to correctly label some test set points when referred to do so by the model.~\cite{kivlichan2021} propose the use of predictive uncertainty as the review criterion so that the test instances with higher uncertainty are sent to the reviewer. Typically, uncertainty is measured by predictive entropy. Figure~\ref{fig:sel_pred} illustrates the results for this task on several datasets, showing the variability of  collaborative accuracy (or ROC-AUC) w.r.t. the fraction of the test set data sent to the oracle for review. We observe that the proposed approach is advantageous in this setting, outperforming the corresponding feature extractor. This shows the usefulness of the predictive uncertainty characterization capability of our proposed approach. 
 



\section{Conclusion}
\label{sec:conclusion}
In this paper, we present a novel uncertainty-aware learning framework for graph-level learning tasks and apply it to graph-level classification and regression. Our approach is general; it can incorporate various existing GNNs/graph transformers and various models from the neural process and meta-learning literature. Experimental results show that various instantiations of the proposed framework offer improved prediction performance on several benchmark tasks, outperform existing uncertainty characterization techniques such as MC dropout, and emerge as a strong competitor to the ensemble with considerably fewer learnable parameters and lower training time.

\bibliographystyle{icml2023}
\bibliography{references}

\begin{thebibliography}{65}
\providecommand{\natexlab}[1]{#1}
\providecommand{\url}[1]{\texttt{#1}}
\expandafter\ifx\csname urlstyle\endcsname\relax
  \providecommand{\doi}[1]{doi: #1}\else
  \providecommand{\doi}{doi: \begingroup \urlstyle{rm}\Url}\fi

\bibitem[Baek et~al.(2021)Baek, Kang, and Hwang]{baek2021}
Baek, J., Kang, M., and Hwang, S.~J.
\newblock Accurate learning of graph representations with graph multiset pooling.
\newblock In \emph{Proc. Int. Conf. Learning Representations}, 2021.

\bibitem[Bianchi et~al.(2020)Bianchi, Grattarola, and Alippi]{bianchi2020mincutpool}
Bianchi, F.~M., Grattarola, D., and Alippi, C.
\newblock Spectral clustering with graph neural networks for graph pooling.
\newblock In \emph{Proc. Int. Conf. Machine learning}, pp.\  2729--2738, 2020.

\bibitem[{Carr} \& {Wingate}(2019){Carr} and {Wingate}]{carr2019}
{Carr}, A. and {Wingate}, D.
\newblock Graph neural processes: {T}owards {B}ayesian graph neural networks.
\newblock \emph{ArXiv e-prints, arXiv:1902.10042}, 2019.

\bibitem[{Day} et~al.(2020){Day}, {Cangea}, {Jamasb}, and {Li{\`o}}]{day2020}
{Day}, B., {Cangea}, C., {Jamasb}, A.~R., and {Li{\`o}}, P.
\newblock Message passing neural processes.
\newblock \emph{ArXiv e-prints: arXiv 2009.13895}, 2020.

\bibitem[Defferrard et~al.(2016)Defferrard, Bresson, and Vandergheynst]{defferrard2016}
Defferrard, M., Bresson, X., and Vandergheynst, P.
\newblock Convolutional neural networks on graphs with fast localized spectral filtering.
\newblock In \emph{Proc. Adv. Neural Info. Process. Syst.}, pp.\  3844--3852, Barcelona, Spain, Dec. 2016.

\bibitem[Duval \& Malliaros(2022)Duval and Malliaros]{duval2022}
Duval, A. and Malliaros, F.
\newblock Higher-order clustering and pooling for graph neural networks.
\newblock In \emph{Proc. ACM Int. Conf. Info. \& Knowl. Management}, pp.\  426–435, 2022.

\bibitem[Duvenaud et~al.(2015)Duvenaud, Maclaurin, et~al.]{duvenaud2015}
Duvenaud, D., Maclaurin, D., et~al.
\newblock Convolutional networks on graphs for learning molecular fingerprints.
\newblock In \emph{Proc. Adv. Neural Inf. Proc. Systems}, 2015.

\bibitem[Dwivedi et~al.(2020)Dwivedi, Joshi, Luu, Laurent, Bengio, and Bresson]{dwivedi2020benchmarkgnns}
Dwivedi, V.~P., Joshi, C.~K., Luu, A.~T., Laurent, T., Bengio, Y., and Bresson, X.
\newblock Benchmarking graph neural networks.
\newblock \emph{arXiv preprint arXiv:2003.00982}, 2020.

\bibitem[Dwivedi et~al.(2021)Dwivedi, Luu, Laurent, Bengio, and Bresson]{dwivedi2021graph}
Dwivedi, V.~P., Luu, A.~T., Laurent, T., Bengio, Y., and Bresson, X.
\newblock Graph neural networks with learnable structural and positional representations.
\newblock \emph{arXiv preprint arXiv:2110.07875}, 2021.

\bibitem[Elinas et~al.(2020)Elinas, Bonilla, and Tiao]{elinas2020}
Elinas, P., Bonilla, E.~V., and Tiao, L.
\newblock {V}ariational inference for graph convolutional networks in the absence of graph data and adversarial settings.
\newblock In \emph{Proc. Adv. Neural Info. Process. Syst.}, volume~33, pp.\  18648--18660, Virtual, Dec. 2020.

\bibitem[Gal \& Ghahramani(2016)Gal and Ghahramani]{gal2016}
Gal, Y. and Ghahramani, Z.
\newblock Dropout as a {B}ayesian approximation: {R}epresenting model uncertainty in deep learning.
\newblock In \emph{Proc. Int. Conf. Machine Learning}, 2016.

\bibitem[Gao \& Ji(2019)Gao and Ji]{gao2019}
Gao, H. and Ji, S.
\newblock Graph u-nets.
\newblock In \emph{Proc. Int. Conf. Machine Learning}, 2019.

\bibitem[{Garnelo} et~al.(2018{\natexlab{a}}){Garnelo}, {Rosenbaum}, {Maddison}, {Ramalho}, {Saxton}, {Shanahan}, {Whye Teh}, {Rezende}, and {Eslami}]{garnelo2018}
{Garnelo}, M., {Rosenbaum}, D., {Maddison}, C.~J., {Ramalho}, T., {Saxton}, D., {Shanahan}, M., {Whye Teh}, Y., {Rezende}, D.~J., and {Eslami}, S.~M.~A.
\newblock {Conditional Neural Processes}.
\newblock \emph{arXiv e-prints}, art. arXiv:1807.01613, July 2018{\natexlab{a}}.

\bibitem[{Garnelo} et~al.(2018{\natexlab{b}}){Garnelo}, {Schwarz}, {Rosenbaum}, {Viola}, {Rezende}, {Eslami}, and {Whye Teh}]{garnelo2018b}
{Garnelo}, M., {Schwarz}, J., {Rosenbaum}, D., {Viola}, F., {Rezende}, D.~J., {Eslami}, S.~M.~A., and {Whye Teh}, Y.
\newblock {Neural Processes}.
\newblock \emph{arXiv e-prints}, art. arXiv:1807.01622, July 2018{\natexlab{b}}.

\bibitem[Gilmer et~al.(2017)Gilmer, Schoenholz, Riley, Vinyals, and Dahl]{gilmer2017neural}
Gilmer, J., Schoenholz, S.~S., Riley, P.~F., Vinyals, O., and Dahl, G.~E.
\newblock Neural message passing for quantum chemistry.
\newblock In \emph{Proc. Int. Conf. Mach. Learn.}, pp.\  1263--1272. PMLR, 2017.

\bibitem[Gordon et~al.(2019)Gordon, Bronskill, Bauer, Nowozin, and Turner]{gordon2018metalearning}
Gordon, J., Bronskill, J., Bauer, M., Nowozin, S., and Turner, R.
\newblock Meta-learning probabilistic inference for prediction.
\newblock In \emph{Proc. Int. Conf. Learning Representations}, 2019.

\bibitem[Griffa et~al.(2017)Griffa, Ricaud, Benzi, Bresson, Daducci, Vandergheynst, Thiran, and Hagmann]{griffa2017transient}
Griffa, A., Ricaud, B., Benzi, K., Bresson, X., Daducci, A., Vandergheynst, P., Thiran, J.-P., and Hagmann, P.
\newblock Transient networks of spatio-temporal connectivity map communication pathways in brain functional systems.
\newblock \emph{NeuroImage}, 155:\penalty0 490--502, 2017.

\bibitem[Hakhamaneshi et~al.(2022)Hakhamaneshi, Nassar, Phielipp, Abbeel, and Stojanovic]{hakhamaneshi2022pretraining}
Hakhamaneshi, K., Nassar, M., Phielipp, M., Abbeel, P., and Stojanovic, V.
\newblock Pretraining graph neural networks for few-shot analog circuit modeling and design.
\newblock \emph{IEEE Trans. Comput.-Aided Des. Integr. Circuits Syst}, 2022.

\bibitem[Hasanzadeh et~al.(2020)Hasanzadeh, Hajiramezanali, Boluki, Zhou, Duffield, Narayanan, and Qian]{hasanzadeh2020}
Hasanzadeh, A., Hajiramezanali, E., Boluki, S., Zhou, M., Duffield, N., Narayanan, K., and Qian, X.
\newblock Bayesian graph neural networks with adaptive connection sampling.
\newblock In \emph{Proc. Int. Conf. Machine Learning}, pp.\  4094--4104, Virtual, Jul. 2020.

\bibitem[Hern{\'a}ndez-Lobato \& Adams(2015)Hern{\'a}ndez-Lobato and Adams]{hernandez2015}
Hern{\'a}ndez-Lobato, J.~M. and Adams, R.
\newblock Probabilistic backpropagation for scalable learning of {B}ayesian neural networks.
\newblock In \emph{Proc. Int. Conf. Machine Learning}, 2015.

\bibitem[Hu et~al.(2020)Hu, Liu*, Gomes, Zitnik, Liang, Pande, and Leskovec]{hu2020StrategiesPretrainingGraph}
Hu, W., Liu*, B., Gomes, J., Zitnik, M., Liang, P., Pande, V., and Leskovec, J.
\newblock Strategies for {{Pre-training Graph Neural Networks}}.
\newblock In \emph{Proc. {{Int}}. {{Conf}}. {{Learn}}. {{Representations}}}, March 2020.

\bibitem[Hu et~al.(2021)Hu, Fey, Zitnik, Dong, Ren, Liu, Catasta, and Leskovec]{hu2021OpenGraphBenchmark}
Hu, W., Fey, M., Zitnik, M., Dong, Y., Ren, H., Liu, B., Catasta, M., and Leskovec, J.
\newblock Open {{Graph Benchmark}}: {{Datasets}} for {{Machine Learning}} on {{Graphs}}.
\newblock In \emph{Proc. Adv. {{Neural Inf}}. {{Process}}. {{Syst}}.}, February 2021.

\bibitem[Irwin et~al.(2012)Irwin, Sterling, Mysinger, Bolstad, and Coleman]{irwin2012ZINCFreeTool}
Irwin, J.~J., Sterling, T., Mysinger, M.~M., Bolstad, E.~S., and Coleman, R.~G.
\newblock {{ZINC}}: {{A Free Tool}} to {{Discover Chemistry}} for {{Biology}}.
\newblock \emph{J. Chem. Inf. Model.}, 52\penalty0 (7):\penalty0 1757--1768, July 2012.
\newblock ISSN 1549-9596.
\newblock \doi{10.1021/ci3001277}.

\bibitem[{Izmailov} et~al.(2021){Izmailov}, {Vikram}, {Hoffman}, and {Wilson}]{izmailov2021}
{Izmailov}, P., {Vikram}, S., {Hoffman}, M.~D., and {Wilson}, A.~G.
\newblock What are {B}ayesian neural network posteriors really like?
\newblock In \emph{Proc. Int. Conf. Machine Learning}, Virtual, Jul. 2021.

\bibitem[{Kim} et~al.(2019){Kim}, {Mnih}, {Schwarz}, {Garnelo}, {Eslami}, {Rosenbaum}, {Vinyals}, and {Whye Teh}]{kim2019}
{Kim}, H., {Mnih}, A., {Schwarz}, J., {Garnelo}, M., {Eslami}, A., {Rosenbaum}, D., {Vinyals}, O., and {Whye Teh}, Y.
\newblock Attentive neural processes.
\newblock In \emph{Proc. Int. Conf. Learning Representations}, January 2019.

\bibitem[Kipf \& Welling(2017)Kipf and Welling]{kipf2017}
Kipf, T. and Welling, M.
\newblock Semi-supervised classification with graph convolutional networks.
\newblock In \emph{Proc. Int. Conf. Learning Representations}, Toulon, France, Apr. 2017.

\bibitem[Kivlichan et~al.(2021)Kivlichan, Liu, Vasserman, and Lin]{kivlichan2021}
Kivlichan, I., Liu, J., Vasserman, L.~H., and Lin, Z. (eds.).
\newblock \emph{Measuring and Improving Model-Moderator Collaboration using Uncertainty Estimation}, 2021.

\bibitem[Korattikara et~al.(2015)Korattikara, Rathod, Murphy, and Welling]{korattikara2015}
Korattikara, A., Rathod, V., Murphy, K., and Welling, M.
\newblock Bayesian dark knowledge.
\newblock In \emph{Proc. Adv. Neural Info. Process. Syst.}, Montreal, Canada, Dec. 2015.

\bibitem[Kreuzer et~al.(2021)Kreuzer, Beaini, Hamilton, L{\'e}tourneau, and Tossou]{kreuzer2021rethinking}
Kreuzer, D., Beaini, D., Hamilton, W., L{\'e}tourneau, V., and Tossou, P.
\newblock Rethinking graph transformers with spectral attention.
\newblock In \emph{Proc. Adv. {{Neural Inf}}. {{Process}}. {{Syst}}.}, volume~34, pp.\  21618--21629, 2021.

\bibitem[Lakshminarayanan et~al.(2017)Lakshminarayanan, Pritzel, and Blundell]{lakshminarayanan2017ensemble}
Lakshminarayanan, B., Pritzel, A., and Blundell, C.
\newblock Simple and scalable predictive uncertainty estimation using deep ensembles.
\newblock In \emph{Proc. Adv. Neural Info. Process. Syst.}, pp.\  6405–6416, 2017.

\bibitem[Lee et~al.(2020{\natexlab{a}})Lee, Hong, and Kim]{lee2020rnp}
Lee, B.-J., Hong, S., and Kim, K.-E.
\newblock Residual neural processes.
\newblock In \emph{Proc. AAAI Conf. Artificial Intell.}, volume~34, pp.\  4545--4552, 2020{\natexlab{a}}.

\bibitem[Lee et~al.(2019)Lee, Lee, and Kang]{lee2019}
Lee, J., Lee, I., and Kang, J.
\newblock Self-attention graph pooling.
\newblock In \emph{Proc. Int. Conf. Machine Learning}, 2019.

\bibitem[Lee et~al.(2020{\natexlab{b}})Lee, Lee, Kim, Yang, Hwang, and Teh]{lee2020bnp}
Lee, J., Lee, Y., Kim, J., Yang, E., Hwang, S.~J., and Teh, Y.~W.
\newblock Bootstrapping neural processes.
\newblock In \emph{Proc. Adv. Neural Inf. Process. Syst.}, 2020{\natexlab{b}}.

\bibitem[Li et~al.(2016)Li, Chen, Carlson, and Carin]{li2016}
Li, C., Chen, C., Carlson, D., and Carin, L.
\newblock Pre-conditioned stochastic gradient {L}angevin dynamics for deep neural networks.
\newblock In \emph{Proc. AAAI Conf. Artificial Intell.}, pp.\  1788–1794, Phoenix, ARI, USA, Feb. 2016.

\bibitem[Li et~al.(2017)Li, Jamieson, DeSalvo, Rostamizadeh, and Talwalkar]{li2017hyperband}
Li, L., Jamieson, K., DeSalvo, G., Rostamizadeh, A., and Talwalkar, A.
\newblock Hyperband: A novel bandit-based approach to hyperparameter optimization.
\newblock \emph{J. Mach. Learn. Res.}, 18\penalty0 (1):\penalty0 6765--6816, 2017.

\bibitem[Liaw et~al.(2018)Liaw, Liang, Nishihara, Moritz, Gonzalez, and Stoica]{liaw2018tune}
Liaw, R., Liang, E., Nishihara, R., Moritz, P., Gonzalez, J.~E., and Stoica, I.
\newblock Tune: A research platform for distributed model selection and training.
\newblock \emph{arXiv preprint arXiv:1807.05118}, 2018.

\bibitem[Liu et~al.(2020)Liu, Li, Chen, Hu, and Huang]{liu2020}
Liu, Z.-Y., Li, S.-Y., Chen, S., Hu, Y., and Huang, S.-J.
\newblock Uncertainty aware graph {G}aussian process for semi-supervised learning.
\newblock In \emph{Proc. AAAI Conf. Artificial Intell.}, pp.\  4957--4964, 2020.

\bibitem[Louizos et~al.(2017)Louizos, Ullrich, and Welling]{louizos2017b}
Louizos, C., Ullrich, K., and Welling, M.
\newblock Bayesian compression for deep learning.
\newblock In \emph{Proc. Adv. Neural Info. Process. Syst.}, pp.\  3288--3298, Long Beach, CA, USA, Dec. 2017.

\bibitem[Ma et~al.(2019)Ma, Tang, Zhu, and Mei]{ma2019}
Ma, J., Tang, W., Zhu, J., and Mei, Q.
\newblock A flexible generative framework for graph-based semi-supervised learning.
\newblock In \emph{Proc. Adv. Neural Info. Process. Syst.}, pp.\  3276--3285, Vancouver, Canada, Dec. 2019.

\bibitem[Monti et~al.(2017)Monti, Bronstein, and Bresson]{monti2017geometric}
Monti, F., Bronstein, M., and Bresson, X.
\newblock Geometric matrix completion with recurrent multi-graph neural networks.
\newblock In \emph{Adv. Neural Inf. Process. Syst.}, volume~30, 2017.

\bibitem[Monti et~al.(2019)Monti, Frasca, Eynard, Mannion, and Bronstein]{monti2019fake}
Monti, F., Frasca, F., Eynard, D., Mannion, D., and Bronstein, M.~M.
\newblock Fake news detection on social media using geometric deep learning.
\newblock \emph{arXiv preprint arXiv:1902.06673}, 2019.

\bibitem[Morris et~al.(2019)Morris, Ritzert, Fey, Hamilton, Lenssen, Rattan, and Grohe]{morris2019weisfeiler}
Morris, C., Ritzert, M., Fey, M., Hamilton, W.~L., Lenssen, J.~E., Rattan, G., and Grohe, M.
\newblock Weisfeiler and leman go neural: Higher-order graph neural networks.
\newblock In \emph{Proc. AAAI Conf. Artif. Intell.}, pp.\  4602--4609, 2019.

\bibitem[Morris et~al.(2020)Morris, Kriege, Bause, Kersting, Mutzel, and Neumann]{morris2020tudataset}
Morris, C., Kriege, N.~M., Bause, F., Kersting, K., Mutzel, P., and Neumann, M.
\newblock Tudataset: A collection of benchmark datasets for learning with graphs.
\newblock In \emph{Proc. Int. Conf. Mach. Learn. Graph Representation Learn. Beyond Workshop}, 2020.

\bibitem[{M{\"u}ller} et~al.(2021){M{\"u}ller}, {Hollmann}, {Pineda Arango}, {Grabocka}, and {Hutter}]{muller2021}
{M{\"u}ller}, S., {Hollmann}, N., {Pineda Arango}, S., {Grabocka}, J., and {Hutter}, F.
\newblock {Transformers Can Do Bayesian Inference}.
\newblock \emph{arXiv e-prints}, December 2021.

\bibitem[Neal(1993)]{neal1993}
Neal, R.~M.
\newblock Bayesian learning via stochastic dynamics.
\newblock In \emph{Proc. Adv. Neural Inf. Proc. Systems}, pp.\  475--482, 1993.

\bibitem[Ng et~al.(2018)Ng, Colombo, and Silva]{ng2018}
Ng, Y.~C., Colombo, N., and Silva, R.
\newblock Bayesian semi-supervised learning with graph gaussian processes.
\newblock In \emph{Proc. Adv. Neural Inf. Process. Syst.}, 2018.

\bibitem[Nguyen \& Grover(2022)Nguyen and Grover]{nguyen2022tnp}
Nguyen, T. and Grover, A.
\newblock Transformer neural processes: Uncertainty-aware meta learning via sequence modeling.
\newblock In \emph{Proc. Int. Conf. Machine Learning}, 2022.

\bibitem[Nogales(2022)]{nogales2022}
Nogales, A.~G.
\newblock On bayesian estimation of densities and sampling distributions: The posterior predictive distribution as the bayes estimator.
\newblock \emph{Stat. Neerl.}, 76\penalty0 (2):\penalty0 236--250, 2022.

\bibitem[Opolka \& Li\'o(2022)Opolka and Li\'o]{opolka2022}
Opolka, F. and Li\'o, P.
\newblock Bayesian link prediction with deep graph convolutional gaussian processes.
\newblock In \emph{Proc. Int. Conf. Artificial Intell. and Statist.}, pp.\  4835--4852, 2022.

\bibitem[Pal et~al.(2020)Pal, Malekmohammadi, Regol, Zhang, Xu, and Coates]{pal2020}
Pal, S., Malekmohammadi, S., Regol, F., Zhang, Y., Xu, Y., and Coates, M.
\newblock Non-parametric graph learning for {B}ayesian graph neural networks.
\newblock In \emph{Proc. Conf. Uncertainty in Artificial Intell.}, Virtual, Aug. 2020.

\bibitem[Ramp{\'a}{\v{s}}ek et~al.(2022)Ramp{\'a}{\v{s}}ek, Galkin, Dwivedi, Luu, Wolf, and Beaini]{rampavsek2022recipe}
Ramp{\'a}{\v{s}}ek, L., Galkin, M., Dwivedi, V.~P., Luu, A.~T., Wolf, G., and Beaini, D.
\newblock Recipe for a general, powerful, scalable graph transformer.
\newblock \emph{arXiv preprint arXiv:2205.12454}, 2022.

\bibitem[Ranjan et~al.(2020)Ranjan, Sanyal, and Talukdar]{ranjan2020}
Ranjan, E., Sanyal, S., and Talukdar, P.~P.
\newblock Asap: {A}daptive structure aware pooling for learning hierarchical graph representations.
\newblock In \emph{Proc. AAAI Conf. Artificial Intell.}, 2020.

\bibitem[Snell et~al.(2017)Snell, Swersky, and Zemel]{snell2017proto}
Snell, J., Swersky, K., and Zemel, R.
\newblock Prototypical networks for few-shot learning.
\newblock In \emph{Proc. Adv. Neural Inf. Process. Syst.}, 2017.

\bibitem[Sun et~al.(2020)Sun, Guo, Zhang, Zhang, Regol, Hu, Guo, Tang, Yuan, He, and Coates]{sun2020}
Sun, J., Guo, W., Zhang, D., Zhang, Y., Regol, F., Hu, Y., Guo, H., Tang, R., Yuan, H., He, X., and Coates, M.
\newblock A framework for recommending accurate and diverse items using {B}ayesian graph convolutional neural networks.
\newblock In \emph{Proc. ACM SIGKDD Int. Conf. Knowl. Discov. \& Data Mining}, pp.\  2030–2039, Virtual, Aug. 2020.

\bibitem[Sun et~al.(2017)Sun, Chen, and Carin]{sun2017}
Sun, S., Chen, C., and Carin, L.
\newblock Learning structured weight uncertainty in {B}ayesian neural networks.
\newblock In \emph{Proc. Int. Conf. Artificial Intell. and Statist.}, pp.\  1283--1292, Ft. Lauderdale, FL, USA, Apr. 2017.

\bibitem[Vaswani et~al.(2017)Vaswani, Shazeer, Parmar, Uszkoreit, Jones, Gomez, Kaiser, and Polosukhin]{vaswani2017attention}
Vaswani, A., Shazeer, N., Parmar, N., Uszkoreit, J., Jones, L., Gomez, A.~N., Kaiser, {\L}., and Polosukhin, I.
\newblock Attention is all you need.
\newblock In \emph{Proc. Adv. Neural Inf. Process. Syst.}, 2017.

\bibitem[Wang et~al.(2020)Wang, Li, Ma, Mont\'{u}far, Zhuang, and Fan]{wang2020haar}
Wang, Y.~G., Li, M., Ma, Z., Mont\'{u}far, G., Zhuang, X., and Fan, Y.
\newblock Haar graph pooling.
\newblock In \emph{Proc. Int. Conf. Machine Learning}, 2020.

\bibitem[Wu et~al.(2022)Wu, Chen, Xu, and Li]{wu2022}
Wu, J., Chen, X., Xu, K., and Li, S.
\newblock Structural entropy guided graph hierarchical pooling.
\newblock In \emph{Proc. Int. Conf. Machine Learning}, Jul. 2022.

\bibitem[Xu et~al.(2019)Xu, Hu, Leskovec, and Jegelka]{xu2019}
Xu, K., Hu, W., Leskovec, J., and Jegelka, S.
\newblock How powerful are graph neural networks?
\newblock In \emph{Proc. Int. Conf. Learning Representations}, New Orleans, LA, USA, May 2019.

\bibitem[Yang et~al.(2022)Yang, Xia, and Chu]{yang2022_LS_prediction}
Yang, C., Xia, Y., and Chu, Z.
\newblock The prediction of the quality of results in logic synthesis using transformer and graph neural networks.
\newblock \emph{arXiv preprint arXiv:2207.11437}, 2022.

\bibitem[{Ying} et~al.(2021){Ying}, {Cai}, {Luo}, {Zheng}, {Ke}, {He}, {Shen}, and {Liu}]{ying2021}
{Ying}, C., {Cai}, T., {Luo}, S., {Zheng}, S., {Ke}, G., {He}, D., {Shen}, Y., and {Liu}, T.-Y.
\newblock Do transformers really perform bad for graph representation?
\newblock In \emph{Proc. Adv. Neural Inf. Process. Syst.}, 2021.

\bibitem[Ying et~al.(2018{\natexlab{a}})Ying, He, Chen, Eksombatchai, Hamilton, and Leskovec]{ying2018graph}
Ying, R., He, R., Chen, K., Eksombatchai, P., Hamilton, W.~L., and Leskovec, J.
\newblock Graph convolutional neural networks for web-scale recommender systems.
\newblock In \emph{Proc. ACM SIGKDD Int. Conf. Knowl Discov. Data Min. (KDD)}, pp.\  974--983, 2018{\natexlab{a}}.

\bibitem[Ying et~al.(2018{\natexlab{b}})Ying, You, Morris, Ren, Hamilton, and Leskovec]{ying2018diffpool}
Ying, Z., You, J., Morris, C., Ren, X., Hamilton, W., and Leskovec, J.
\newblock Hierarchical graph representation learning with differentiable pooling.
\newblock In \emph{Proc. Adv. Neural Inf. Process. Syst.}, volume~31, 2018{\natexlab{b}}.

\bibitem[Zhang et~al.(2018)Zhang, Cui, Neumann, and Chen]{zhang2018}
Zhang, M., Cui, Z., Neumann, M., and Chen, Y.
\newblock An end-to-end deep learning architecture for graph classification.
\newblock In \emph{Proc. AAAI Conf. Artificial Intell.}, 2018.

\bibitem[Zhang et~al.(2019)Zhang, Pal, Coates, and {\"U}stebay]{zhang2019}
Zhang, Y., Pal, S., Coates, M., and {\"U}stebay, D.
\newblock Bayesian graph convolutional neural networks for semi-supervised classification.
\newblock In \emph{Proc. AAAI Conf. Artificial Intell.}, pp.\  5829--5836, Feb. 2019.

\end{thebibliography}

\newpage
\appendix
\onecolumn

\section{Experimental Details}

\subsection{Details of Datasets}
\label{sec:app:dataset}
Table.~\ref{tab:dataset_stat} summarizes the statistics of the graph datasets used for the experiments in this paper.

\begin{table*}[h!]
\centering
\caption{Summary statistics of the graph datasets~\cite{morris2020tudataset, hu2021OpenGraphBenchmark, dwivedi2020benchmarkgnns}}
\footnotesize
\label{tab:dataset_stat}
\setlength{\tabcolsep}{2pt}
\begin{tabular}{l|cccccccccc}
\toprule
 TU Datasets &D\&D         &PROTEINS     &MUTAG        &NCI1       &PTC       &IMDB-B     &IMDB-M        &REDDIT-B     &REDDIT-M       &COLLAB      \\ \midrule
\# Graphs & 1178 &  1113 &  188 &  4110 & 344 & 1000 & 1500 & 2000 & 5000  & 5000 \\  
Avg. \# nodes &  284.32 &  39.06 & 17.93 & 29.8 & 25.5 & 19.8 & 13.0 & 429.6 & 508.5 & 74.5 \\
\# classes & 2 & 2 &  2 & 2 & 2 & 2 & 3 & 2 & 5 & 3 \\
Metric &  \multicolumn{10}{c}{Accuracy} \\
\midrule
 OGB Datasets &  BACE         &  BBBP &  CLINTOX & HIV  &  SIDER & TOXCAST     &  TOX21 &    \multicolumn{2}{|l|}{Regression Dataset}  & ZINC    \\ \midrule 
\# Graphs &   1513 & 2039 & 1478 &   41127 & 1427 & 8575 & 7831 & \multicolumn{2}{|l|}{\# Graphs} & 12000   \\  
Avg. \# nodes &  33.6 & 22.5 & 25.5 & 25.3 &  30.0 &  16.7 & 16.5  & \multicolumn{2}{|l|}{Avg. \# nodes} & 23.2  \\
\# binary classif. tasks & 1 & 1  &  2 & 1 & 27 & 617 & 12 & \multicolumn{2}{|l|}{\# regression tasks} & 1 \\
Metric &  \multicolumn{7}{c}{ROC-AUC}  &  \multicolumn{2}{|l|}{Metric}  & MAE \\
\bottomrule
\end{tabular}
\end{table*}

\paragraph{TU Datasets~\cite{morris2020tudataset}: }

The TU Benchmark consists of social networks  and bioinformatic datasets.

\emph{Social networks datasets:  }  
IMDB-BINARY and IMDB-MULTI are movie collaboration datasets. 
Each graph is the ego-network for each actor/actress, where nodes correspond to actors/actresses and an edge indicates a collaboration between two actors/actresses (i.e., appear in the same movie).
Each graph is derived from a pre-specified genre of movies, 
The task is to predict these genres from the graphs, in a multi-class classification setting.
REDDIT-BINARY and REDDIT-MULTI5K are balanced datasets where each graph corresponds to an online discussion thread on Reddit.
The nodes correspond to users in the online discussion and  
an edge is drawn between two nodes if at least one of them responds to the other’s comments.
The task is to find a mapping from each graph to its community or a `subreddit'. 
COLLAB is derived from 3 public collaboration datasets; namely, High Energy Physics, Condensed Matter Physics and Astro-Physics. 
Each graph is built as an ego-network of different researchers from different fields, and the task is to classify the research area of each author.

\emph{Bioinformatics datasets:  }  
MUTAG is a dataset of 188 mutagenic aromatic and heteroaromatic nitro compounds with 7 discrete labels.
In the PROTEINS dataset, nodes are secondary structure elements (SSEs), whereas the edge between two nodes indicates that they are neighbors in the amino-acid sequence or in 3D space. 
PTC is a dataset of 344 chemical compounds related to carcinogenicity for male and female rats.
NCI1 is publicly available from the National Cancer Institute (NCI) and is a subset of balanced datasets of chemical compounds screened for the ability to suppress or inhibit the growth of a panel of human tumor-cell lines.

\paragraph{OGB Datasets~\cite{hu2021OpenGraphBenchmark}}

The open graph benchmark (OGB) is one of the widely used graph benchmarks, initially proposed in~\cite{hu2021OpenGraphBenchmark}. Many new datasets are included in recent years~\cite{hu2020StrategiesPretrainingGraph}.

BBBP is related to blood-brain barrier penetration (membrane permeability).
TOX21 is from toxicity data on 12 biological targets, including nuclear receptors and stress response pathways.
TOXCAST is derived from the toxicology measurements based on over 600 in-vitro high-throughput screenings.
SIDER is a database of marketed drugs and adverse drug reactions (ADR), grouped into 27 system organ classes.
CLINTOX is qualitative classifying drugs approved by the FDA and those that have failed clinical trials for toxicity reasons.
HIV is constructed based on the experimentally measured abilities to inhibit HIV replication.
BACE is obtined from qualitative binding results for a set of inhibitors of human $\beta$-secretase 1.

\paragraph{ZINC-12K Dataset~\cite{dwivedi2020benchmarkgnns}}

ZINC-12K is a molecular dataset with 12K graphs introduced in \cite{dwivedi2020benchmarkgnns}, which is the subset of the ZINC-full~\cite{irwin2012ZINCFreeTool}.
The task is to predict the constrained solubility (logP) of the molecule. This dataset has a predefined 10K/1K/1K train/validation/test split.

\subsection{Hyperparameters}
\label{sec:app:hyperparam}

For a fair comparison to the graph encoder models, we fixed their hyperparameters, adopting the configurations from the original papers~\cite{baek2021,dwivedi2020benchmarkgnns,rampavsek2022recipe}

We conducted a hyperparameter search for the \textit{PPD} modules.
For the TU and OGB datasets, we tuned the hyperparameter via a grid search while on the ZINC datset, we used the hyperband algorithm~\cite{li2017hyperband} on the \textit{Ray.Tune} platform~\cite{liaw2018tune}.

We provide a summary of the hyperparameter configurations in Table~\ref{tab:hparam}.

\begin{table*}[h!]
\centering
\caption{List of hyperparameter configurations for GIN/GMT (GINE/GraphGPS) (One number if sharing the same hyperparameter)}
\footnotesize
\label{tab:hparam}
\setlength{\tabcolsep}{2pt}
\begin{tabular}{l|cccccccccc}
\toprule
GIN &D\&D         &PROTEINS     &MUTAG        &NCI1       &PTC       &IMDB-B     &IMDB-M        &REDDIT-B     &REDDIT-M       &COLLAB      \\ \midrule
\# enc. layer & 1 & 2 &  1 & 1/2  &  2 & 2 & 1/2 & 1/2 & 1 & 1 
\\
\# dec. layer & 1 & 2 &  2   & 2/2  & 1/2 & 2 & 1 &  1 & 2/1 & 1
 \\ 
\# attn. layer &  1 & 1/2 & 1  &  1/2 & 2/1 & 1/2 & 2/1 &  2 &  2/1 & 1
\\
\# attn. head & 1 &  2/1  &  1  & 2 &   1  & 2/4 & 1 & 1/2 & 2/4 & 1
\\
Batch Size &  10 & 128/128 &  128 & 128  & 128 & 128 &  128  & 128 & 128 & 128
\\
Context Size & 64 & 64 & 64  & 64 &  & 64 & 64 & 64 & 64 &  64 
 \\ 
\midrule
 OGB Datasets &  BACE         &  BBBP &  CLINTOX & HIV  &  SIDER & TOXCAST     &  TOX21 &    \multicolumn{2}{|l|}{Regression Dataset}  & ZINC    \\ \midrule 
\# enc. layer & 2 &  1 & 1 & 1 &  2 &  1 & 1 &  \multicolumn{2}{|l|}{\# enc. layer}  & 2
\\
\# dec. layer & 1 & 2 & 1 &  1 & 2 & 2 & 2 & \multicolumn{2}{|l|}{ \# dec. layer }  & 1
\\
\# attn. layer & 1 & 2 &  1 &  2 & 2 & 1 & 1 & \multicolumn{2}{|l|}{\# attn. layer} & 2
\\
\# attn. head &  2 & 4 &  1 &  1 & 1 & 4 & 2 & \multicolumn{2}{|l|}{\# attn. head } & 2
\\
Batch Size &  128 & 128  & 128 & 512 &  128 & 128 & 128 & \multicolumn{2}{|l|}{Batch Size} & 32
\\
Context Size &  64 &  64 &  64 & 64 &  64 & 64 & 64 & \multicolumn{2}{|l|}{Context Size} & 32
\\
\bottomrule
\end{tabular}
\end{table*}

\subsection{Detailed Experimental Results}
\label{sec:app:detailed_res}
The detailed results including performance metrics, training time, and number of learnable parameters from our experiments on OGB, TU, and ZINC-12k datasets are shown in Tables~\ref{tab:results_ogb_base},~\ref{tab:results_tu_base}, and~\ref{tab:results_zinc_base}, respectively.

\begin{table*}[ht]
\centering
\caption{Comparison of ROC-AUC of molecular property prediction, training time, and no. learnable parameters for OGB datasets. Relative increase and no. win/tie/loss is computed w.r.t. the corresponding graph encoder model in each case.}
\scriptsize
\label{tab:results_ogb_base}
\setlength{\tabcolsep}{2pt}
\begin{tabular}{l|ccccccc|cc}
\hline
Alg. &BACE         &BBBP     &CLINTOX        &HIV &SIDER &TOXCAST &TOX21  &rel. increase &\#(win/tie/loss)    \\ \hline \hline
GINE  
 &73.5$\pm$2.0 &68.0$\pm$1.6 &86.0$\pm$2.4 &75.7$\pm$2.0 &55.3$\pm$1.9 &60.9$\pm$0.5 &73.2$\pm$0.9 &- &-\\
GINE-MC 
 &73.4$\pm$2.1 &67.9$\pm$1.6 &86.3$\pm$2.7 &75.6$\pm$2.0 &55.3$\pm$1.7 &60.8$\pm$0.5 &73.1$\pm$1.0  &-0.02\% &1/1/5\\
GINE-Ensmb.   &74.4$\pm$1.8 &69.0$\pm$1.1 &88.3$\pm$1.8 &78.0$\pm$0.7 &56.8$\pm$1.5 &61.8$\pm$0.5 &73.8$\pm$0.4 &1.91\% &7/0/0 \\
Ours (2-stage)   &71.4$\pm$6.0 &68.2$\pm$1.7 &87.3$\pm$2.2 &75.9$\pm$2.1 &56.0$\pm$1.2 &61.6$\pm$0.5 &73.6$\pm$0.8 &0.33\% &6/0/1\\
Ours (E2E)   &74.2$\pm$3.2 &69.8$\pm$2.1 &86.7$\pm$2.4 &75.0$\pm$1.3 &57.8$\pm$1.1 &61.9$\pm$0.7 &74.1$\pm$0.8 &1.59\% &6/0/1\\ \hline
GMT    &67.9$\pm$6.8 &67.3$\pm$0.8 &82.3$\pm$5.5 &76.1$\pm$2.1 &53.4$\pm$3.1 &63.8$\pm$0.6 &75.7$\pm$0.6 &- &-\\
GMT-MC  &67.9$\pm$6.7 &67.3$\pm$0.8 &83.2$\pm$4.4 &76.2$\pm$2.1 &53.4$\pm$2.9 &63.8$\pm$0.6 &75.9$\pm$0.6 &0.21\%  &2/4/1\\ 
GMT-Ensmb.  &68.7$\pm$4.5 &68.2$\pm$1.4 &86.2$\pm$3.5 &77.1$\pm$1.0 &55.1$\pm$2.5 &64.3$\pm$0.6 &76.6$\pm$0.2 &1.95\% &6/0/1\\
Ours(2-stage)  &71.4$\pm$9.7 &67.6$\pm$1.3 &82.1$\pm$2.2 &76.5$\pm$1.6 &55.3$\pm$2.8 &64.9$\pm$0.9 &76.3$\pm$0.7 &1.70\% &6/0/1 \\
Ours(E2E) &76.6$\pm$5.4 &68.1$\pm$1.4 &85.5$\pm$2.9 &75.5$\pm$1.0 &56.8$\pm$1.8 &66.2$\pm$0.7 &75.8$\pm$0.4 &3.93\% &6/0/1\\
\hline 
\hline
Training time (Sec.) & & & & & & &  \\
\hline \hline
GINE  &160.4 &188.1 &154.2 &1654.4 &161.6 &1110.9 &708.7 &- &-\\
GINE-MC &160.4 &188.1 &154.2 &1654.4 &161.6 &1110.9 &708.7 &0.0\% &-\\
GINE-Ensmb. &320.8 &376.1 &308.3 &3308.9 &323.1 &2221.7 &1417.4 &100.0\% &-\\
Ours (2-stage)  &257.7 &314.8 &238.9 &2844.5 &270.6 &1999.3 &1192.9 &67.2\% &-\\
Ours (E2E) &201.2 &247.1 &204.0 &2037.4 &215.9 &1380.1 &978.1 &29.7\%\\    \hline
GMT   &214.3 &242.2 &198.2 &2122.9 &209.8 &1355.0 &942.5 &-\\
GMT-MC    &214.3 &242.2 &198.2 &2122.9 &209.8 &1355.0 &942.5 &0.0\% &-\\
GMT-Ensmb. &428.6 &484.4 &396.4 &4245.9 &419.6 &2710.1 &1885.1 &100.0\% &-\\
Ours (2-stage)    &324.7 &398.5 &311.0 &3488.8 &334.5 &2479.8 &1472.3 &62.3\% &-
 \\
Ours (E2E)   &248.6 &325.7 &250.7 &2283.6 &257.8 &1757.2 &1241.8  &24.1\% &-\\
\hline 
\hline
No. parameters ($\times 10^6$) & & & & & & & \\
\hline \hline
GINE  &0.301 &0.301 &0.301 &0.301 &0.302 &0.341 &0.301  &-\\
GINE-MC  &0.301 &0.301 &0.301 &0.301 &0.302 &0.341 &0.301 &0.0\% &-\\
GINE-Ensmb.  &0.601 &0.601 &0.601 &0.601 &0.604 &0.681 &0.603 &100.0\% &-\\
Ours (2-stage) &0.474 &0.474 &0.375 &0.441 &0.497 &0.566 &0.477 &53.7\% &-\\
Ours (E2E) &0.474 &0.474 &0.375 &0.441 &0.497 &0.566 &0.477 &53.7\% &-\\
\hline
GMT    &0.946 &0.946 &0.946 &0.943 &0.948 &0.982 &0.943 &- &-\\
GMT-MC &0.946 &0.946 &0.946 &0.943 &0.948 &0.982 &0.943 &0.0\% &-\\
GMT-Ensmb.  &1.893 &1.891 &1.891 &1.886 &1.896 &1.964 &1.886 &100.0\% &-\\
Ours (2-stage)  &1.037 &1.102 &1.037 &1.067 &1.077 &1.336 &1.052 &15.7\% &-\\
Ours (E2E)   &1.037 &1.102 &1.037 &1.067 &1.077 &1.336 &1.052 &15.7\% &-\\
\hline \hline
\end{tabular}
\end{table*}

\begin{table*}[ht]
\centering
\caption{Comparison of accuracy, training time, and number of learnable parameters for different models for TU datasets. Relative increase and no. win/tie/loss is computed w.r.t. the corresponding graph encoder model in each case.}
\scriptsize
\label{tab:results_tu_base}
\setlength{\tabcolsep}{1.5pt}
\begin{tabular}{l|ccccc|ccccc|cc}
\hline
Accuracy &D\&D         &PROTEINS     &MUTAG        &NCI1       &PTC       &IMDB-B     &IMDB-M        &REDDIT-B     &REDDIT-M       &COLLAB  &rel. increase &\#(win/tie/loss)    \\ \hline \hline
GIN    &72.1$\pm$1.5 &72.0$\pm$1.2 &83.4$\pm$1.5 &78.3$\pm$0.5 &54.6$\pm$1.4 &74.3$\pm$0.5 &51.2$\pm$0.4 &91.5$\pm$0.4 &55.6$\pm$0.6 &81.1$\pm$0.4 &- &- \\
GIN-MC  &72.3$\pm$1.5 &71.6$\pm$1.4 &83.4$\pm$1.7 &78.1$\pm$0.5 &54.5$\pm$1.3 &74.2$\pm$0.5 &51.2$\pm$0.5 &91.5$\pm$0.3 &55.5$\pm$0.6 &81.2$\pm$0.3 &-0.10\%  &2/3/5\\
GIN-Ensmb.
&75.0$\pm$0.7 &73.1$\pm$0.8 &83.5$\pm$1.1 &80.3$\pm$0.4 &55.0$\pm$1.2 &74.1$\pm$0.6 &51.7$\pm$0.4 &91.6$\pm$0.2 &56.1$\pm$0.3 &81.6$\pm$0.2 &1.13\% &9/0/1 \\
Ours (2-stage) 
&74.3$\pm$0.9 &74.9$\pm$0.6 &84.8$\pm$1.0 &80.0$\pm$0.4 &56.5$\pm$1.8 &73.6$\pm$0.7 &51.1$\pm$0.7 &91.8$\pm$0.3 &56.5$\pm$0.4 &81.3$\pm$0.4 &1.54\%  &8/0/2 \\
Ours (E2E)
&72.2$\pm$1.4 &72.0$\pm$1.3 &84.0$\pm$3.3 &78.7$\pm$0.4 &54.7$\pm$1.6 &74.3$\pm$0.7 &51.2$\pm$0.6 &91.8$\pm$0.8 &56.5$\pm$0.3 &81.7$\pm$0.4 &0.44\%
  &7/2/1\\
\hline
GMT    &78.3$\pm$0.5 &74.8$\pm$0.9 &82.7$\pm$0.6 &76.3$\pm$0.4 &56.0$\pm$2.7 &73.7$\pm$0.8 &50.6$\pm$0.5 &91.9$\pm$0.2 &55.7$\pm$0.3 &80.4$\pm$0.3 &- &- \\
GMT-MC  &78.3$\pm$0.5 &74.9$\pm$0.7 &82.6$\pm$0.7 &76.3$\pm$0.4 &56.2$\pm$3.0 &73.8$\pm$0.7 &50.5$\pm$0.7 &91.9$\pm$0.2 &55.7$\pm$0.3 &80.5$\pm$0.3 &0.06\%  &4/4/2\\
GMT-Ensmb.  &78.8$\pm$0.2 &75.5$\pm$0.6 &82.8$\pm$0.8 &76.8$\pm$0.2 &56.6$\pm$1.3 &74.1$\pm$0.6 &51.2$\pm$0.3 &92.0$\pm$0.2 &55.9$\pm$0.2 &81.3$\pm$0.1 &0.68\%  &10/0/0\\
Ours (2-stage) 
&78.5$\pm$0.8 &75.4$\pm$0.5 &85.2$\pm$0.7 &76.9$\pm$0.4 &55.7$\pm$1.8 &73.5$\pm$0.6 &51.2$\pm$0.3 &92.3$\pm$0.2 &55.8$\pm$0.2 &81.1$\pm$0.3 &0.69\% &8/0/2 \\
Ours (E2E)
&78.4$\pm$0.4 &74.9$\pm$0.6 &81.3$\pm$1.1 &76.6$\pm$0.4 &57.2$\pm$1.7 &73.5$\pm$0.8 &50.1$\pm$0.9 &91.9$\pm$0.6 
&56.1$\pm$0.5 &81.3$\pm$0.3 &0.19\%  &6/1/3\\
\hline \hline
ECE & & & & & & & & & & \\
\hline \hline
GIN   &0.128$\pm$0.009 &0.134$\pm$0.008 &0.169$\pm$0.008 &0.070$\pm$0.004 &0.233$\pm$0.021 &0.122$\pm$0.007 &0.109$\pm$0.005 &0.053$\pm$0.003 &0.078$\pm$0.003 &0.065$\pm$0.003 
& &- \\
GIN-MC  &0.121$\pm$0.006 &0.132$\pm$0.009 &0.174$\pm$0.013 &0.066$\pm$0.004 &0.210$\pm$0.019 &0.119$\pm$0.009 &0.109$\pm$0.004 &0.057$\pm$0.002 &0.076$\pm$0.003 &0.059$\pm$0.004 
& &7/1/2 \\
GIN-Ensmb.
&0.122$\pm$0.007 &0.132$\pm$0.006 &0.166$\pm$0.008 &0.067$\pm$0.005 &0.225$\pm$0.015 &0.119$\pm$0.005 &0.102$\pm$0.006 &0.056$\pm$0.003 &0.077$\pm$0.004 &0.058$\pm$0.003 
& &9/0/1 \\
Ours (2-stage) 
&0.175$\pm$0.006 &0.126$\pm$0.005 &0.153$\pm$0.010 &0.071$\pm$0.003 &0.286$\pm$0.023 &0.134$\pm$0.009 &0.126$\pm$0.003 &0.052$\pm$0.002 &0.071$\pm$0.003 &0.069$\pm$0.003 
& &4/0/6 \\
Ours (E2E)
&0.180$\pm$0.012 &0.136$\pm$0.008 &0.156$\pm$0.019 &0.071$\pm$0.003 &0.251$\pm$0.025 &0.130$\pm$0.005 &0.118$\pm$0.008 &0.051$\pm$0.003 &0.072$\pm$0.004 &0.061$\pm$0.002 
& &5/0/5\\
\hline
GMT    &0.113$\pm$0.005 &0.126$\pm$0.006 &0.196$\pm$0.012 &0.072$\pm$0.004 &0.196$\pm$0.015 &0.127$\pm$0.010 &0.116$\pm$0.007 &0.055$\pm$0.002 &0.075$\pm$0.003 &0.059$\pm$0.003 
& &- \\
GMT-MC  &0.110$\pm$0.007 &0.130$\pm$0.008 &0.202$\pm$0.015 &0.068$\pm$0.004 &0.192$\pm$0.016 &0.122$\pm$0.010 &0.113$\pm$0.006 &0.055$\pm$0.003 &0.077$\pm$0.003 &0.056$\pm$0.003 
& &6/1/3\\
GMT-Ensmb.  &0.111$\pm$0.004 &0.133$\pm$0.006 &0.187$\pm$0.013 &0.065$\pm$0.005 &0.186$\pm$0.016 &0.125$\pm$0.007 &0.114$\pm$0.005 &0.054$\pm$0.003 &0.073$\pm$0.002 &0.056$\pm$0.003 & &9/0/1\\
Ours (2-stage) 
&0.116$\pm$0.006 &0.128$\pm$0.007 &0.172$\pm$0.010 &0.072$\pm$0.005 &0.207$\pm$0.015 &0.127$\pm$0.005 &0.130$\pm$0.005 &0.051$\pm$0.003 &0.077$\pm$0.005 &0.065$\pm$0.003 
& &2/2/6\\
Ours (E2E)
&0.116$\pm$0.006 &0.128$\pm$0.006 &0.199$\pm$0.017 &0.071$\pm$0.004 &0.227$\pm$0.017 &0.136$\pm$0.008 &0.121$\pm$0.005 &0.053$\pm$0.003 &0.071$\pm$0.003 &0.063$\pm$0.005 
& &3/0/7 \\
\hline \hline
NLL & & & & & & & & & & \\
\hline \hline
GIN   &0.416$\pm$0.016 &0.456$\pm$0.015 &0.416$\pm$0.058 &0.367$\pm$0.005 &0.556$\pm$0.026 &0.357$\pm$0.006 &0.466$\pm$0.002 &0.165$\pm$0.005 &0.308$\pm$0.002 &0.228$\pm$0.003 
& &- \\
GIN-MC  &0.391$\pm$0.006 &0.418$\pm$0.013 &0.304$\pm$0.028 &0.352$\pm$0.004 &0.524$\pm$0.016 &0.356$\pm$0.005 &0.465$\pm$0.001 &0.166$\pm$0.004 &0.308$\pm$0.002 &0.219$\pm$0.003 
& &8/1/1 \\
GIN-Ensmb.
&0.376$\pm$0.006 &0.441$\pm$0.019 &0.328$\pm$0.041 &0.336$\pm$0.002 &0.518$\pm$0.012 &0.349$\pm$0.002 &0.461$\pm$0.002 &0.159$\pm$0.002 &0.304$\pm$0.001 &0.216$\pm$0.003  
& &10/0/0 \\
Ours (2-stage) 
&0.577$\pm$0.043 &0.427$\pm$0.018 &0.353$\pm$0.063 &0.350$\pm$0.004 &0.644$\pm$0.043 &0.374$\pm$0.008 &0.474$\pm$0.003 &0.160$\pm$0.004 &0.301$\pm$0.001 &0.223$\pm$0.003 
& &6/0/4 \\
Ours (E2E)
&0.585$\pm$0.036 &0.469$\pm$0.020 &0.359$\pm$0.033 &0.359$\pm$0.003 &0.585$\pm$0.045 &0.362$\pm$0.005 &0.467$\pm$0.003 &0.157$\pm$0.006 &0.299$\pm$0.001 &0.210$\pm$0.003 
& &5/0/5 \\
\hline
GMT    &0.351$\pm$0.005 &0.387$\pm$0.010 &0.316$\pm$0.020 &0.372$\pm$0.003 &0.520$\pm$0.019 &0.362$\pm$0.006 &0.472$\pm$0.005 &0.168$\pm$0.005 &0.308$\pm$0.001 &0.213$\pm$0.002  
& &- \\
GMT-MC  &0.345$\pm$0.004 &0.382$\pm$0.008 &0.309$\pm$0.016 &0.369$\pm$0.003 &0.512$\pm$0.015 &0.359$\pm$0.005 &0.470$\pm$0.004 &0.160$\pm$0.004 &0.308$\pm$0.001 &0.209$\pm$0.002  
& &9/0/1 \\
GMT-Ensmb.  &0.345$\pm$0.004 &0.377$\pm$0.003 &0.305$\pm$0.014 &0.363$\pm$0.002 &0.501$\pm$0.008 &0.352$\pm$0.002 &0.467$\pm$0.002 &0.163$\pm$0.003 &0.307$\pm$0.001 &0.204$\pm$0.001
& &10/0/0 \\
Ours (2-stage) 
&0.353$\pm$0.005 &0.393$\pm$0.015 &0.294$\pm$0.015 &0.369$\pm$0.003 &0.531$\pm$0.017 &0.363$\pm$0.006 &0.478$\pm$0.003 &0.157$\pm$0.003 &0.307$\pm$0.001 &0.211$\pm$0.003 
& &5/0/5 \\
Ours (E2E)
&0.349$\pm$0.007 &0.386$\pm$0.007 &0.337$\pm$0.015 &0.371$\pm$0.004 &0.534$\pm$0.016 &0.372$\pm$0.013 &0.472$\pm$0.003 &0.157$\pm$0.005 &0.303$\pm$0.002 &0.208$\pm$0.003  
& &6/1/3 \\
\hline \hline
Brier Score & & & & & & & & & & \\
\hline \hline
GIN   &0.416$\pm$0.016 &0.456$\pm$0.015 &0.416$\pm$0.058 &0.367$\pm$0.005 &0.556$\pm$0.026 &0.357$\pm$0.006 &0.466$\pm$0.002 &0.165$\pm$0.005 &0.308$\pm$0.002 &0.228$\pm$0.003 
& &- \\
GIN-MC  &0.391$\pm$0.006 &0.418$\pm$0.013 &0.304$\pm$0.028 &0.352$\pm$0.004 &0.524$\pm$0.016 &0.356$\pm$0.005 &0.465$\pm$0.001 &0.166$\pm$0.004 &0.308$\pm$0.002 &0.219$\pm$0.003 
& &8/1/1 \\
GIN-Ensmb.
&0.376$\pm$0.006 &0.441$\pm$0.019 &0.328$\pm$0.041 &0.336$\pm$0.002 &0.518$\pm$0.012 &0.349$\pm$0.002 &0.461$\pm$0.002 &0.159$\pm$0.002 &0.304$\pm$0.001 &0.216$\pm$0.003  
& &10/0/0 \\
Ours (2-stage) 
&0.577$\pm$0.043 &0.427$\pm$0.018 &0.353$\pm$0.063 &0.350$\pm$0.004 &0.644$\pm$0.043 &0.374$\pm$0.008 &0.474$\pm$0.003 &0.160$\pm$0.004 &0.301$\pm$0.001 &0.223$\pm$0.003 
& &6/0/4 \\
Ours (E2E)
&0.585$\pm$0.036 &0.469$\pm$0.020 &0.359$\pm$0.033 &0.359$\pm$0.003 &0.585$\pm$0.045 &0.362$\pm$0.005 &0.467$\pm$0.003 &0.157$\pm$0.006 &0.299$\pm$0.001 &0.210$\pm$0.003  
& &5/0/5\\
\hline
GMT    &0.351$\pm$0.005 &0.387$\pm$0.010 &0.316$\pm$0.020 &0.372$\pm$0.003 &0.520$\pm$0.019 &0.362$\pm$0.006 &0.472$\pm$0.005 &0.168$\pm$0.005 &0.308$\pm$0.001 &0.213$\pm$0.002 
& &- \\
GMT-MC  &0.345$\pm$0.004 &0.382$\pm$0.008 &0.309$\pm$0.016 &0.369$\pm$0.003 &0.512$\pm$0.015 &0.359$\pm$0.005 &0.470$\pm$0.004 &0.160$\pm$0.004 &0.308$\pm$0.001 &0.209$\pm$0.002   
& &9/1/0\\
GMT-Ensmb.  &0.345$\pm$0.004 &0.377$\pm$0.003 &0.305$\pm$0.014 &0.363$\pm$0.002 &0.501$\pm$0.008 &0.352$\pm$0.002 &0.467$\pm$0.002 &0.163$\pm$0.003 &0.307$\pm$0.001 &0.204$\pm$0.001 
& &10/0/0 \\
Ours (2-stage) 
&0.353$\pm$0.005 &0.393$\pm$0.015 &0.294$\pm$0.015 &0.369$\pm$0.003 &0.531$\pm$0.017 &0.363$\pm$0.006 &0.478$\pm$0.003 &0.157$\pm$0.003 &0.307$\pm$0.001 &0.211$\pm$0.003 
& &5/0/5 \\
Ours (E2E)
&0.349$\pm$0.007 &0.386$\pm$0.007 &0.337$\pm$0.015 &0.371$\pm$0.004 &0.534$\pm$0.016 &0.372$\pm$0.013 &0.472$\pm$0.003 &0.157$\pm$0.005 &0.303$\pm$0.002 &0.208$\pm$0.003   
& &6/1/3 \\
\hline \hline
Training time (Sec.) & & & & & & & & & & \\
\hline \hline
GIN     &67.4 &12.9 &3.3 &47.5 &3.2 &11.4 &18.6 &115.5 &283.0 &149.4 &- &-\\
GIN-MC   &67.4 &12.9 &3.3 &47.5 &3.2 &11.4 &18.6 &115.5 &283.0 &149.4 &0.0\% &-\\
GIN-Ensmb.
&202.1 &25.9 &10.0 &142.5 &9.5 &22.9 &37.2 &346.6 &849.1 &298.8 &160.2\% &-\\
Ours(2-stage) 
 &228.1 &23.8 &7.1 &97.1 &9.5 &30.4 &45.4 &196.2 &485.5 &358.9 &133.2\% &-\\
Ours(E2E)
 &137.5 &13.0 &4.9 &53.2 &4.8 &16.4 &26.6 &186.4 &546.0 &264.3 &53.3\% &-
 \\
\hline
GMT    &169.3 &34.8 &7.1 &166.9 &9.4 &21.9 &31.0 &272.0 &955.6 &258.2 &- &-\\
GMT-MC   &169.3 &34.8 &7.1 &166.9 &9.4 &21.9 &31.0 &272.0 &955.6 &258.2 &0.0\% &-\\
GMT-Ensmb. &338.5 &69.7 &14.3 &333.8 &18.9 &43.8 &62.1 &544.0 &1911.2 &516.3  &100.0\% &-\\
Ours(2-stage) 
 &513.4 &58.4 &14.1 &234.3 &17.4 &48.8 &69.6 &431.1 &1396.4 &601.7 &98.0\% &-\\
Ours(2-stage)
 &361.3 &47.8 &10.9 &220.0 &11.7 &29.3 &38.3 &389.1 &1085.4 &431.8 &44.2\% &-\\
\hline \hline
No. param. ($\times 10^6$) & & & & & & & & & & \\
\hline \hline
GIN     &0.010 &0.109 &0.109 &0.113 &0.111 &0.126 &0.120 &0.108 &0.109 &0.171  &- &-\\
GIN-MC  &0.010 &0.109 &0.109 &0.113 &0.111 &0.126 &0.120 &0.108 &0.109 &0.171 &0.0\% &-
 \\
GIN-Ensmb.    &0.030 &0.218 &0.328 &0.339 &0.332 &0.252 &0.240 &0.325 &0.326 &0.343  &160.0\% &-\\
Ours(2-stage) 
 &0.022 &0.216 &0.233 &0.286 &0.268 &0.233 &0.211 &0.282 &0.267 &0.329 &118.9\% &-\\
Ours (E2E)
&0.022 &0.216 &0.233 &0.286 &0.268 &0.233 &0.211 &0.282 &0.267 &0.329 &118.9\% &-
 \\
\hline
GMT     &0.066 &0.901 &0.902 &0.907 &0.903 &0.916 &0.909 &0.938 &0.946 &0.967  &-\\
GMT-MC  &0.066 &0.901 &0.902 &0.907 &0.903 &0.916 &0.909 &0.938 &0.946 &0.967 &0.0\%\\
GMT-Ensmb. &0.133 &1.802 &1.804 &1.814 &1.806 &1.832 &1.819 &1.876 &1.892 &1.934 &100.0\%\\
Ours (2-stage) 
&0.071 &0.976 &1.092 &1.080 &1.076 &1.106 &1.050 &1.128 &1.104 &1.091 &16.1\% \\
Ours (E2E)
&0.071 &0.976 &1.092 &1.080 &1.076 &1.106 &1.050 &1.128 &1.104 &1.091 &16.1\% \\
\hline \hline
\end{tabular}
\end{table*}

\begin{table}[ht]
\centering
\caption{Mean and standard error of MAE ($\downarrow$) for ZINC-12k dataset}
\footnotesize
\label{tab:results_zinc_base}
\setlength{\tabcolsep}{3pt}
\begin{tabular}{l|ccc|ccc}
\toprule
Alg.  &GINE &Ours(2-stage) &Ours(E2E)  &GraphGPS  &Ours(2-stage) &Ours(E2E) \\ \midrule
MAE $\downarrow$ &0.070$\pm$0.004 &0.067$\pm$0.003 &0.065$\pm$0.004 &0.070$\pm$0.004 
    &0.068$\pm$0.002 
    &0.067$\pm$0.003 \\
\bottomrule
\end{tabular} 
\end{table}

\end{document}